\documentclass[journal,twoside,web]{ieeecolor}
\usepackage{generic}
\usepackage{cite}
\usepackage{amsmath,amssymb,amsfonts}
\usepackage{graphicx}
\usepackage{hyperref}
\hypersetup{hidelinks=true}
\usepackage{textcomp}
\usepackage{balance}
\usepackage{color}
\usepackage{makecell}
\usepackage{nicefrac}       
\usepackage{microtype}      
\usepackage[table]{xcolor}         
\usepackage{float}
\usepackage[T1]{fontenc}    
\usepackage{tcolorbox}
\tcbuselibrary{skins, breakable, theorems}
\usepackage{longtable} 
\usepackage{pifont}
\usepackage{xcolor}
\usepackage{CJKutf8}
\usepackage[utf8]{inputenc}
\usepackage{tabulary}
\usepackage{tabularx}
\usepackage{threeparttable}
\usepackage{ltxtable}
\usepackage{booktabs}
\usepackage{multirow}
\usepackage{subfigure}
\usepackage{algpseudocode}

\def\BibTeX{{\rm B\kern-.05em{\sc i\kern-.025em b}\kern-.08em
    T\kern-.1667em\lower.7ex\hbox{E}\kern-.125emX}}
\begin{document}
\title{PREBA: Surgical Duration Prediction via PCA-Weighted Retrieval-Augmented LLMs \\and Bayesian Averaging Aggregation}

\author{Wanyin Wu\textsuperscript{\textdagger}, Kanxue Li\textsuperscript{\textdagger}, Baosheng Yu, Haoyun Zhao, Yibing Zhan, ~\IEEEmembership{Member,~IEEE},\\ Dapeng Tao*, ~\IEEEmembership{Member,~IEEE}, and Hua Jin*
\thanks{Wanyin Wu, Haoyun Zhao, and Dapeng Tao are with the School of Information Science and Engineering, Yunnan University, Kunming 650500, China.}
\thanks{Kanxue Li and Yibing Zhan are with the School of Computer Science, Wuhan University, Wuhan 430072, China.}
\thanks{Baosheng Yu is with the Lee Kong Chian School of Medicine, Nanyang Technological University, Singapore 639798.}
\thanks{Hua Jin is with First People’s Hospital of Yunnan Province, Kunming 650032, China.}
\thanks{\textsuperscript{\textdagger}Wanyin Wu and Kanxue Li contributed equally to this work. Author order is random.}
\thanks{* Corresponding author: Dapeng Tao and Hua Jin}
}

\maketitle

\begin{abstract}

Accurate prediction of surgical duration is pivotal for hospital resource management. Although recent supervised learning approaches, from machine learning (ML) to fine-tuned large language models (LLMs), have shown strong performance, they remain constrained by the need for high-quality labeled data and computationally intensive training. 
In contrast, zero-shot LLM inference offers a promising training-free alternative, but it lacks grounding in institution-specific clinical evidence, making its predictions clinically misaligned and prone to instability. 
To address these limitations, we present PREBA, a retrieval-augmented framework that integrates PCA-weighted retrieval and Bayesian averaging to ground LLM reasoning in institution-specific evidence without requiring task-specific fine-tuning. 
Specifically, PREBA first encodes heterogeneous clinical features into a unified representation. 
It then performs PCA-weighted retrieval, weighting each dimension by its contribution to the variation across institutional cases, instead of treating them as equally informative, and conditions the LLM to generate multiple candidate predictions.
Finally, Bayesian averaging consolidates these predictions with population-level statistical priors, yielding a calibrated estimate that remains stable and clinically plausible.
We evaluate PREBA on four real-world datasets spanning different institutions and languages, using three state-of-the-art LLMs, including Qwen3, DeepSeek-R1, and HuatuoGPT-o1. PREBA consistently outperforms zero-shot inference, reducing MAE by up to 64.9\% and raising R² to as high as 0.63.
It further attains accuracy competitive with, and on several datasets superior to, fully supervised ML methods, while offering greater interpretability through retrieved exemplars and explicit priors. The code is available at https://github.com/kanxueli/PREBA.

\end{abstract}


\section{Introduction}
\label{sec1}
\IEEEPARstart{S}{urgical} duration prediction plays a vital role in hospital management, particularly in optimizing operating-room (OR) utilization and resource allocation~\cite{sdp_manage,sdp_manage1}.
Accurate predictions provide reliable evidence for surgical scheduling, effectively preventing underutilization
of expensive operating room resources, thereby reducing patient waiting times and enhancing care coordination efficiency~\cite{ml_2,svr}. 
However, due to complex clinical data and non-stationary dynamics of surgical workflows, reliable surgical duration prediction remains a significant challenge~\cite{sdp_manage2,azriel2024surgery}.

Surgical duration prediction has evolved from experience-based estimation to increasingly data-driven AI approaches over the past decades~\cite{nn_1,ml_4}. Early approaches relied on surgeons' empirical judgment and simple statistical averages of historical records, but often struggled to capture the multifactorial complexity of surgical procedures~\cite{sdp_manage}. With the accumulation of structured perioperative records, machine learning methods such as Decision Trees~\cite{Decision_Tree}, and XGBoost~\cite{XGBoost} improved prediction accuracy by learning nonlinear relationships between structured clinical features and surgical duration. Despite these advances, such approaches remain constrained by their reliance on large volumes of labeled data and intensive training, and their fixed parameters limit adaptation once clinical distributions shift~\cite{nn2,muep}.   
In parallel, large language models have approached the problem along two distinct paths~\cite{llm4sdp}: fine-tuning on curated surgical data and zero-shot inference. The former still requires retraining on institutional data, whereas the latter departs from this altogether, drawing on the model's pretrained knowledge to generate predictions through structured prompting without any task-specific parameter update.

\begin{figure*}[t]
  \centering
  \centerline{\includegraphics[width=1\linewidth,height=0.41\textheight]{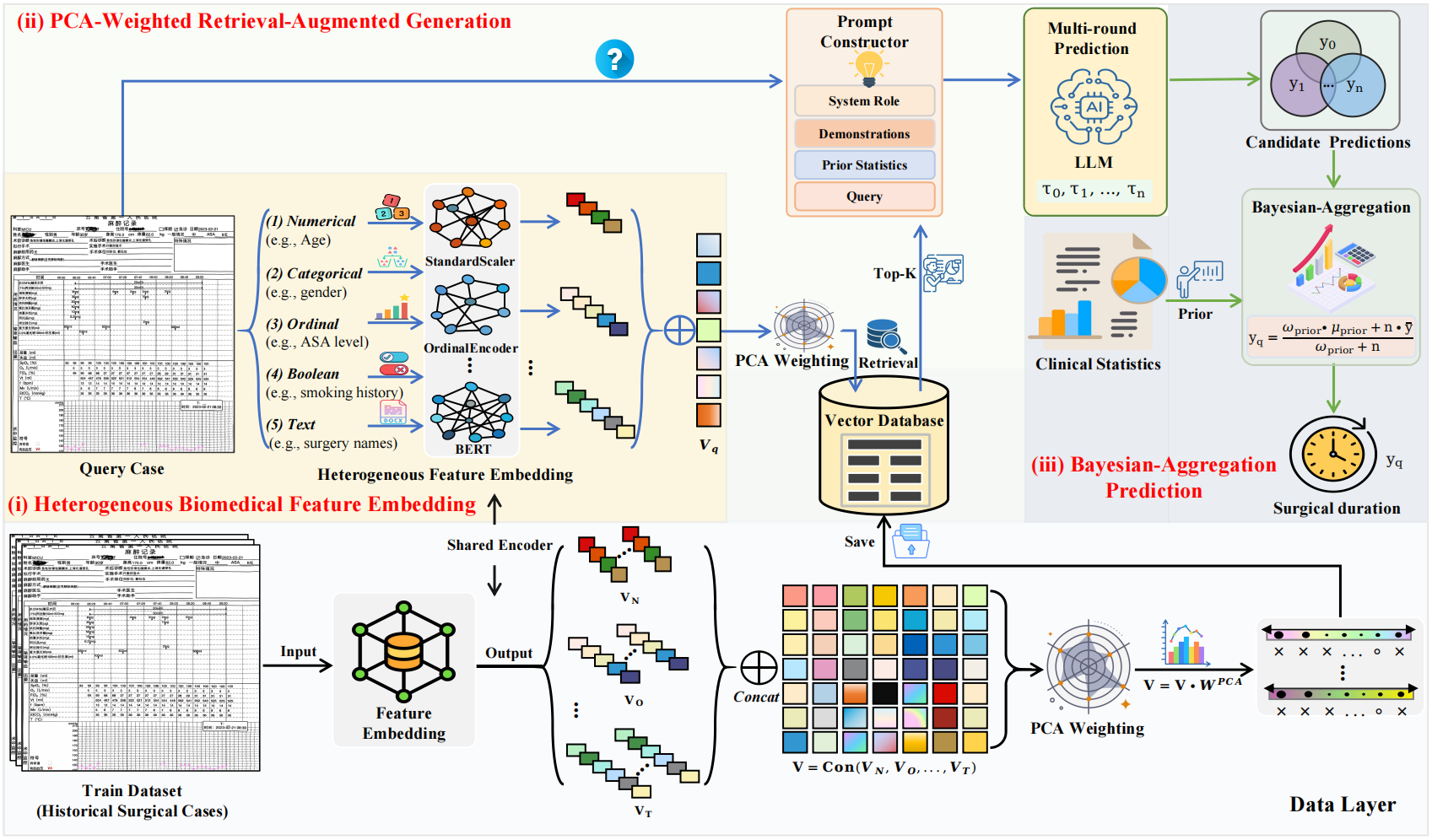}}
  \caption{\textbf{The proposed PREBA framework} for surgical duration prediction, consisting of three key modules: (i) Heterogeneous Biomedical Feature Embedding, (ii) PCA-Weighted Retrieval-Augmented Generation, and (iii) Bayesian Averaging Aggregation.}
  \label{framework}
\end{figure*}

Despite this advantage, zero-shot LLM inference still faces two key limitations that constrain its clinical reliability. First, LLMs are trained on globally diverse medical corpora and therefore tend to encode population-level averages, causing predictions to drift toward generic patterns~\cite{case_mix}. Retrieval-augmented generation (RAG) offers a natural remedy by anchoring predictions in the institution-specific historical cases~\cite{rag_review2}. 
However, retrieving these cases by generic similarity treats all representation dimensions as equally informative, although they differ widely in how much information they carry about how cases differ, and can therefore return references that differ in the factors that actually govern surgical duration~\cite{fang2021patient}.
Second, surgical duration is jointly shaped by multiple interacting clinical factors, and a single inference pass is confined to one particular way of weighing this information, often failing to adequately integrate the full range of relevant clinical considerations. Sampling multiple predictions under varying decoding conditions can alleviate this limitation by exploring different ways of weighing these factors \cite{wan2025reasoning}, yielding an information gain over a single inference pass, but this diversity also introduces a degree of randomness and dispersion across the resulting candidates.

To address these limitations, we propose \textbf{PREBA}, a retrieval-augmented generation framework that integrates \textbf{P}CA-weighted \textbf{RE}trieval and \textbf{B}ayesian \textbf{A}veraging to ground LLM reasoning in institution-specific clinical evidence and population-level priors. As shown in Fig.~\ref{framework}, PREBA first encodes heterogeneous clinical features into a unified representation space. Rather than retrieving cases through generic semantic similarity, PREBA then employs a Principal Component Analysis (PCA)~\cite{pca}-weighted retrieval mechanism that derives weights from the covariance structure of the institution's own historical cases, 
emphasizing the directions that carry most of the variation across the institution’s cases while suppressing low-information dimensions. 
These retrieved cases, together with population-level clinical priors, condition the LLM to generate multiple candidate predictions across repeated sampling rounds. Rather than aggregating these candidates through simple averaging or majority voting~\cite{self_consistency}, PREBA finally applies Bayesian averaging~\cite{bayesian} to calibrate them against institution-specific priors, producing calibrated and clinically plausible estimates.

To validate the efficacy of PREBA, we conduct extensive experiments on four real-world surgical duration datasets using four regression metrics. 
We instantiate PREBA with three state-of-the-art LLMs, Qwen3~\cite{qwen3}, DeepSeek-R1~\cite{deepseek}, and HuatuoGPT-o1~\cite{huatuo-gpt}, comparing it against zero-shot and random few-shot inference under each backbone, as well as nine established ML baselines.
As reported in Table~\ref{tab:main_results}, PREBA consistently outperforms zero-shot and random few-shot inference across all four datasets, reducing MAE by up to 64.9\% relative to zero-shot inference. 
On the in-hospital dataset, for instance, Qwen3-8B's MAE decreases by 40.2\%, from 59.87 to 35.81 minutes, while R\textsuperscript{2} improves from -0.18 to 0.62. 
Notably, PREBA also achieves performance competitive with, and on several datasets superior to, fully supervised ML baselines. As shown in Table~\ref{tab:ml_results}, on MMSDP, HuatuoGPT-o1-7B attains an MAE of 27.85 minutes with R\textsuperscript{2}=0.39, surpassing the strongest ML baseline, XGBoost, at 31.20 minutes. These results demonstrate the effectiveness and generalizability of PREBA for practical deployment in real-world clinical systems.

In summary, our contributions are threefold: 
(1) we propose PREBA, a retrieval-augmented framework that grounds LLM reasoning in institution-specific clinical evidence and population-level priors without task-specific LLM fine-tuning; 
(2) we introduce a PCA-weighted retrieval mechanism derived from institutional data covariance, coupled with a Bayesian averaging strategy that calibrates multi-path predictions against institutional priors, distinguishing PREBA from existing RAG-based and ensemble-aggregation approaches; 
(3) extensive experiments on four real-world datasets with three LLM backbones show that PREBA achieves accuracy competitive with supervised ML models while offering greater interpretability through retrieved exemplars and explicit priors.


\section{Related Work}
\label{sec2}

\subsection{Surgical Duration Prediction Methods}
\label{sec2.2}

Beyond expert judgment and simple statistical averages~\cite{sdp_manage1}, surgical duration has been modeled with tree-based ensembles such as Decision Trees~\cite{Decision_Tree}, Bagging~\cite{Bagging}, and XGBoost~\cite{XGBoost}, as well as Support Vector Regression~\cite{svr} and (quantile) linear models~\cite{QrandForest}. To incorporate unstructured clinical narratives, subsequent work has fused pre-trained language model embeddings, such as BERT, with structured perioperative variables through multimodal encoders~\cite{bert}. More recently, large language models have been applied either by fine-tuning on curated surgical records or through zero-shot prompting that requires no task-specific training~\cite{llm4sdp,luo2024biomedgpt}. Regardless of the modeling paradigm, however, these approaches yield only a static point estimate and offer no explicit mechanism for anchoring predictions in the deploying institution's own case distribution, motivating retrieval-based approaches that ground predictions in institution-specific evidence~\cite{aamodt1994case,li2025cross}.

\subsection{Retrieval-Augmented Generation for Clinical Prediction}
\label{sec2.2}

Retrieval-augmented generation grounds LLM reasoning in external evidence without additional training \cite{rag_review, realtime_tmm, action}, and recent studies have applied this paradigm to clinical prediction by retrieving similar patient records from an institution's own database \cite{medical_rag}. Retrieval in these methods is typically performed via top-$K$ search in a generic semantic space, treating all representation dimensions as equally informative. This is not well suited to surgical records, which mix numerical, categorical, ordinal, and textual attributes, and risks retrieving cases that appear similar in embedding space but differ in the factors that actually govern surgical duration. Supervised alternatives such as ridge coefficients~\cite{guyon2003introduction}, permutation importance~\cite{altmann2010permutation}, or SHAP~\cite{mosca2022shap} could address this, but require duration labels and often an auxiliary fitted model. 
PREBA instead derives retrieval weights from the covariance structure of institutional historical cases via PCA~\cite{pca}, which requires no duration labels.

\subsection{Multi-Path Prediction Aggregation for LLMs}
\label{sec2.2}

Sampling multiple reasoning paths from an LLM under varying decoding temperatures is a common strategy for improving output reliability~\cite{wan2025reasoning,madaan2023self}. This is particularly relevant to surgical duration, which depends on how multiple interacting clinical factors are weighed, and a single inference pass reflects only one such weighing~\cite{sdp_manage1}. Existing approaches typically aggregate the resulting candidates through simple averaging, majority voting, or quantile averaging, following the same principle as self-consistency voting in text generation \cite{self_consistency}.  However, these rules operate solely on the sampled candidates, without reference to the target institution's own historical statistics, so the consolidated estimate can still depart from clinically plausible ranges. PREBA instead aggregates multi-path predictions with institution-specific statistical priors through Bayesian averaging, adapting the prior weight to the amount of matched historical evidence. Overall, PREBA differs from existing RAG-based and multi-path aggregation methods in how retrieval and aggregation are grounded in institution-specific evidence, as summarized in Table~\ref{tab:comparison}.

\setlength{\tabcolsep}{4pt}
\newcolumntype{L}{>{\centering\arraybackslash}m{\dimexpr 0.30\linewidth-2\tabcolsep\relax}}
\newcolumntype{Y}{>{\centering\arraybackslash}m{\dimexpr 0.17\linewidth-2\tabcolsep\relax}}
\newcolumntype{Z}{>{\centering\arraybackslash}m{\dimexpr 0.18\linewidth-2\tabcolsep\relax}}
\begin{table}[t]
\centering
\caption{Comparison of PREBA with Existing Methods.}
\label{tab:comparison}
\footnotesize
\renewcommand{\arraystretch}{1.4}
\rowcolors{2}{gray!12}{white}
\begin{tabular}{@{} L Y Y Z Z @{}}
\toprule
\rowcolor{white}
Dimension & Traditional ML & Fine-tuned LLM & Generic RAG-LLM & PREBA (Ours) \\
\midrule
Requires no parameter retraining    & \texttimes & \texttimes & \checkmark & \checkmark \\
Retrieval weighting           & --         & --         & Uniform    & \textbf{PCA-weighted} \\
Multi-path sampling           & \texttimes & \texttimes & Varies     & Multi-temperature \\
Aggregation of candidates     & --         & --         & Averaging / voting & \textbf{Bayesian averaging} \\
Interpretability              & Limited    & Limited    & Retrieved cases only & \textbf{Cases + priors} \\
\bottomrule
\end{tabular}
\end{table}

\section{Method}\label{sec3}

This section details the proposed PREBA framework for surgical duration prediction. As illustrated in Fig.~\ref{framework}, PREBA consists of three core components. The heterogeneous biomedical feature embedding module first encodes diverse clinical feature types into a unified representation space. Building on this representation, the PCA-weighted retrieval module retrieves clinically relevant historical cases and conditions the LLM to generate multiple candidate predictions. Finally, the Bayesian averaging aggregation module calibrates these candidates against population-level clinical priors, producing stable and clinically plausible duration estimates.



\subsection{Problem Definition}
Surgical duration prediction, defined as estimating the duration from patient entry into the operating room to exit~\cite{sdp_manage}, is formulated as a regression task. Let $D_{\text{his}} = \{(x_i, y_i)\}_{i=1}^N$ denote the historical surgery dataset, where $x_i$ represents the preoperative clinical features of the $i$-th surgery and $y_i \in \mathbb{R}^+$ its true surgical duration in minutes. Given a query case $\mathbf{x}_q \notin \mathcal{D}_{\text{his}}$, the objective is a point estimate $\hat{y}_q \in \mathbb{R}^+$ of its surgical duration. Using $D_{\text{his}}$ as the retrieval corpus, we first identify the top-$K$ most similar historical cases in the PCA-weighted feature space:
\begin{equation}
R_K = \{(x_j, y_j) : j \in \mathcal{N}_K(x_q, D_{\text{his}})\}
\end{equation}
where $\mathcal{N}_K$ denotes the top-$K$ nearest neighbor retrieval function. The query case and retrieved examples are then formulated into a prompt fed to the LLM to generate multiple predictions across $n$ rounds:
\begin{equation}
\hat{y}_q^{(r)} = \text{LLM}(x_q, R_K; \tau_r), \quad r = 1, 2, \ldots, n
\end{equation}
where $\tau_r$ represents the sampling temperature in round $r$. Finally, the $n$ predictions are aggregated with Bayesian averaging~\cite{bayesian} to produce the final estimate $\hat{y}_{q}$:
\begin{equation}
\bar{y} = \frac{1}{n}\sum_{r=1}^{n}\hat{y}_q^{(r)},
\label{Bayesian_eq}
\quad \hat{y}_{q} = \frac{w_{\text{prior}} \cdot \mu_{\text{prior}} + n \cdot \bar{y}}{w_{\text{prior}} + n}
\end{equation}
where $\bar{y}$ denotes the sample mean of LLM predictions, $\mu_{\text{prior}}$ represents the prior mean, and $w_{\text{prior}}$ indicates the prior strength.

\subsection{\textbf{Heterogeneous Biomedical Feature Embedding}}
Clinical electronic health records contain diverse feature modalities spanning multiple data types and representational structures. 
These features originate from distinct stages of the perioperative workflow, including patient demographics and comorbidities recorded at admission, preoperative laboratory and imaging results, planned anesthesia and surgical procedures, and free-text descriptions of the intended operation. 
To enable effective downstream retrieval and generation tasks, these heterogeneous features must be transformed into a unified, dense vector space. 
This subsection details the encoding strategy for constructing this unified representation.

\subsubsection{Type-Specific Feature Encoding}
Clinical features are categorized into five distinct types, each requiring tailored encoding mechanisms. \textit{Numerical features} (e.g., patient age) are standardized via \(\text{StandardScaler}\) to achieve zero mean and unit variance. \textit{Ordinal features} (e.g., surgery level) are encoded via \(\text{OrdinalEncoder}\) to preserve ordering semantics. \textit{Categorical features} (e.g., gender) are processed through \(\text{OneHotEncoder}\) to generate sparse binary representations. \textit{Boolean features} (e.g., abnormal pulmonary function) are encoded via \(\text{LabelEncoder}\) to represent binary conditions. All non-textual encoders are implemented using scikit-learn~\cite{scikit-learn}. \textit{Textual features} (e.g., surgery names) are encoded via BERT~\cite{bert} contextual embeddings to capture semantic content.

\subsubsection{Unified Representation Construction}
Within each feature type, the individual clinical features are encoded separately and concatenated into a type-specific sub-vector; for example, the numerical sub-vector \(v_{\text{num}}\) concatenates the standardized encodings of age, weight, height, and other numerical attributes. The five type-specific sub-vectors are then concatenated into a single representation:
\begin{equation}
v^{(\text{emb})} = [v_{\text{num}}, v_{\text{ord}}, v_{\text{cat}}, v_{\text{bool}}, v_{\text{text}}] \in \mathbb{R}^D,
\end{equation}
where \(D = \sum_c \dim(v_c)\) is the effective total dimensionality, aggregated across all five feature types.



\subsection{\textbf{PCA-Weighted Retrieval-Augmented Generation}}\label{subsec32}

The concatenated embedding $v^{(\text{emb})}$ treats every dimension as equally informative, regardless of how much information it carries about how cases differ. Correlated or near-constant dimensions then influence the similarity as much as those that separate cases, distorting the retrieval ranking. We address this limitation with a data-driven PCA-weighted retrieval method that reweights the retrieval space by how much each dimension contributes to the variation across institutional cases.


\begin{figure*}[t!]
\centering
\centerline{\includegraphics[width=1\linewidth]{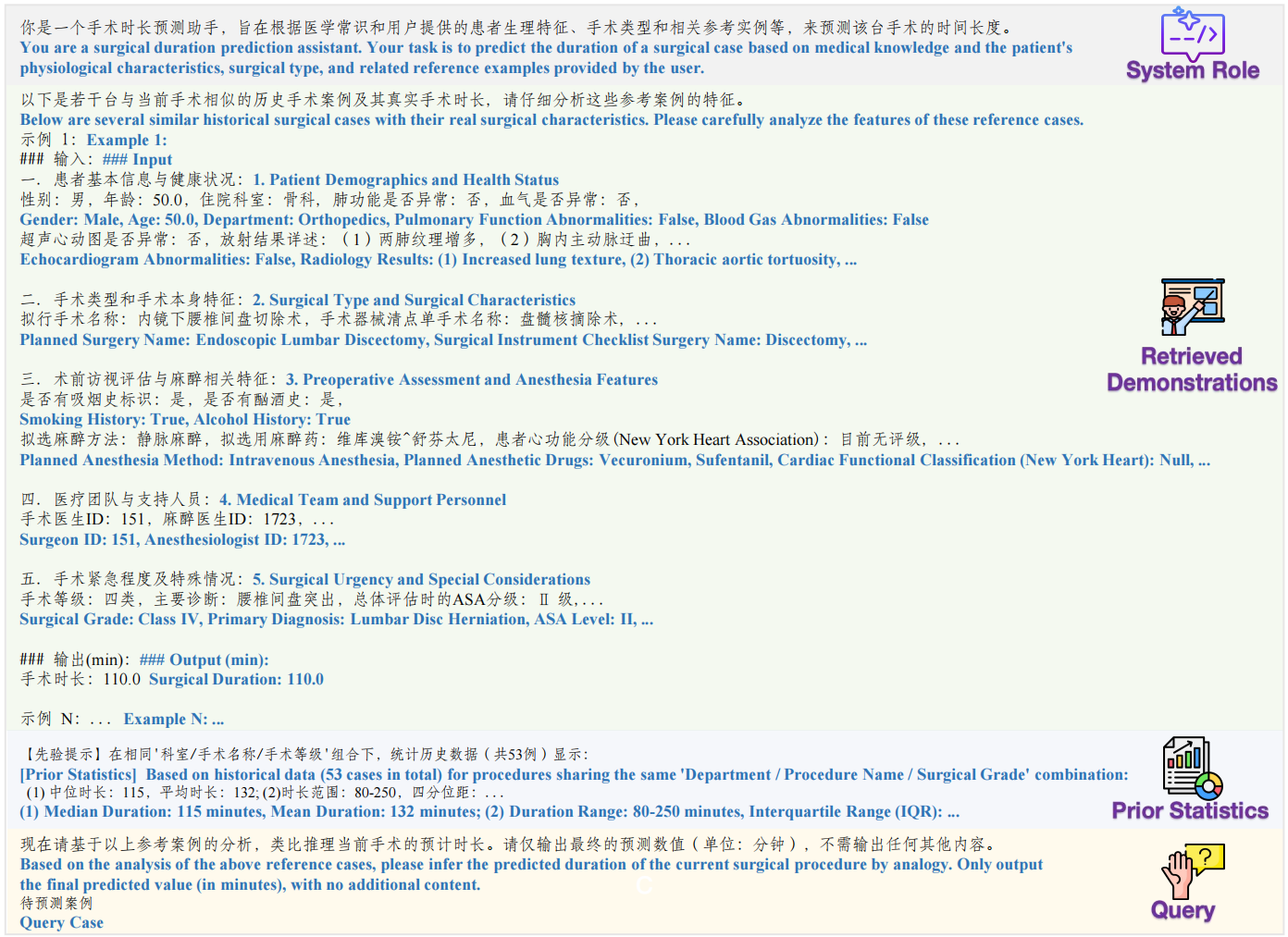}}
\caption{Structured prompt template for retrieval-augmented surgical duration prediction.
The prompt integrates (1) system role definition, (2) similar case demonstrations retrieved via PCA-weighted similarity, (3) statistical priors from historical data, and (4) the query case.}
\label{prompt_template}
\end{figure*}

\subsubsection{Data-Driven Feature Weighting via PCA}
Rather than relying on manually assigned feature weights, we employ Principal Component Analysis (PCA)~\cite{pca} to derive retrieval weights directly from the training data in an unsupervised manner. 
Let $V = [v_1^{(\text{emb})}, v_2^{(\text{emb})}, \ldots, v_N^{(\text{emb})}]^T \in \mathbb{R}^{N \times D}$ denote the matrix of embeddings from $N$ training cases. We perform PCA on the centered data matrix, obtaining the principal components $W = [w_1, w_2, \ldots, w_D] \in \mathbb{R}^{D \times D}$ and the corresponding explained variance ratios $\{\sigma_k\}_{k=1}^{D}$, where $\sigma_k$ is the variance explained by the $k$-th component. The weight for the $j$-th feature dimension aggregates its absolute loadings across the top-$K_{\text{pca}}$ principal components, weighted by their explained variances:
\begin{equation}
w_j^{\text{PCA}} = \frac{1}{K_{\text{pca}}} \sum_{k=1}^{K_{\text{pca}}} |W_{jk}| \cdot \sigma_k
\end{equation}
This assigns higher weights to dimensions that consistently contribute to the directions carrying most of the variation across cases. 
The weighted case embedding is then obtained via element-wise multiplication:
\begin{equation}
v^{(\text{weighted})} = v^{(\text{emb})} \odot w^{\text{PCA}}
\end{equation}
where $w^{\text{PCA}} \in \mathbb{R}^D$ is the PCA-derived weight vector, applied identically to all historical cases and the query.

\subsubsection{Similarity Retrieval with Clinical Post-Processing}
All historical cases are stored in a FAISS~\cite{faiss} vector database as PCA-weighted embeddings $v_i^{(\text{weighted})}$. 
For a query case $\mathbf{x}_q$, we first encode it into $\mathbf{v}_q^{\text{(emb)}}$ using the heterogeneous embedding module, then apply the same PCA weighting to obtain $\mathbf{v}_q^{\text{(weighted)}}$.

The retrieval process begins by computing cosine similarity between the query vector $\mathbf{v}_q^{\text{(weighted)}}$ and all weighted case vectors in the database. The cosine similarity metric~\cite{cosine} (i.e. $\mathcal{N}_K$ function) is defined as:
\begin{equation}
\text{sim}(\mathbf{v}_q, \mathbf{v}_i) = \frac{\mathbf{v}_q \cdot \mathbf{v}_i}{\|\mathbf{v}_q\| \|\mathbf{v}_i\|}
\end{equation}
This metric is chosen for its effectiveness in high-dimensional spaces and its invariance to vector magnitude, focusing purely on directional similarity.
We initially retrieve the top $M = N \cdot K$ most similar cases based on cosine similarity, where $N > 1$ is an expansion factor. This expanded candidate set $\mathcal{C} = \{(\mathbf{x}_i, y_i)\}_{i=1}^M$ ensures sufficient coverage of potentially relevant cases, acknowledging that the raw similarity ranking may not fully align with clinical relevance due to the complexity of surgical contexts.

The initial candidate pool $\mathcal{C}$ is refined via hierarchical filtering to eliminate clinically inconsistent and statistically anomalous cases. First, we restrict candidates to cases from the same medical department as the query, ensuring departmental clinical coherence. Second, we apply hierarchical feature matching based on exact agreement with key clinical attributes (department, planned surgery name, and surgery level) in descending order of specificity. 
Third, we perform statistical outlier removal using interquartile range (IQR)~\cite{iqr} analysis on durations. 
The refined reference set $\mathcal{R}_K$ of size $K$ is selected from the post-processed candidates by highest similarity score.

\subsubsection{Multi-Path LLM Prediction Generation}
The refined reference set $\mathcal{R}_K = \{(\mathbf{x}_i, y_i)\}_{i=1}^K$ and the query case $\mathbf{x}_q$ are synthesized into a structured prompt $\mathcal{P}(\mathbf{x}_q, \mathcal{R}_K, \mathcal{S}_q)$ following the template illustrated in Fig.~\ref{prompt_template}. The prompt construction function $\mathcal{P}$ integrates four key components: (1) a \textit{system prompt} defining the AI's role as a surgical duration prediction assistant; (2) \textit{similar case demonstrations} detailing the clinical features and actual durations of the $K$ reference cases; (3) \textit{statistical prior knowledge} 
$\mathcal{S}_q$ including median, mean, range, and interquartile ranges of surgical durations for the relevant clinical stratum; and (4) the \textit{query case's} clinical profile.
This comprehensive prompt design enables the LLM to engage in analogical reasoning based on specific similar cases while maintaining awareness of population-level statistics. 

Surgical duration depends on how multiple clinical factors, such as procedure complexity, comorbidities, and anesthesia plan, are jointly weighed, yet a single inference pass commits to only one such weighing and cannot reveal how the estimate would change under a different one. We therefore prompt the LLM to generate multiple reasoning paths by sampling $n$ predictions under varying decoding temperatures:
\begin{equation}
\hat{y}_q^{(r)} = \text{LLM}(\mathcal{P}(\mathbf{x}_q, \mathcal{R}_K, \mathcal{S}_q); \tau_r), \quad r = 1, 2, \ldots, n,
\end{equation}
where temperature parameter $\tau_1 = 0$ yields a deterministic prediction and $\tau_r \sim \mathcal{U}(0.05, 0.4)$ for $r \geq 2$ introduces controlled variability across paths. This multi-path sampling strategy yields an ensemble $\{\hat{y}_q^{(1)}, \hat{y}_q^{(2)}, \ldots, \hat{y}_q^{(n)}\}$ that reflects both the model's most confident reasoning and plausible variations in its output space.

\begin{figure*}[t!]
\centering
\centerline{\includegraphics[width=1\linewidth]{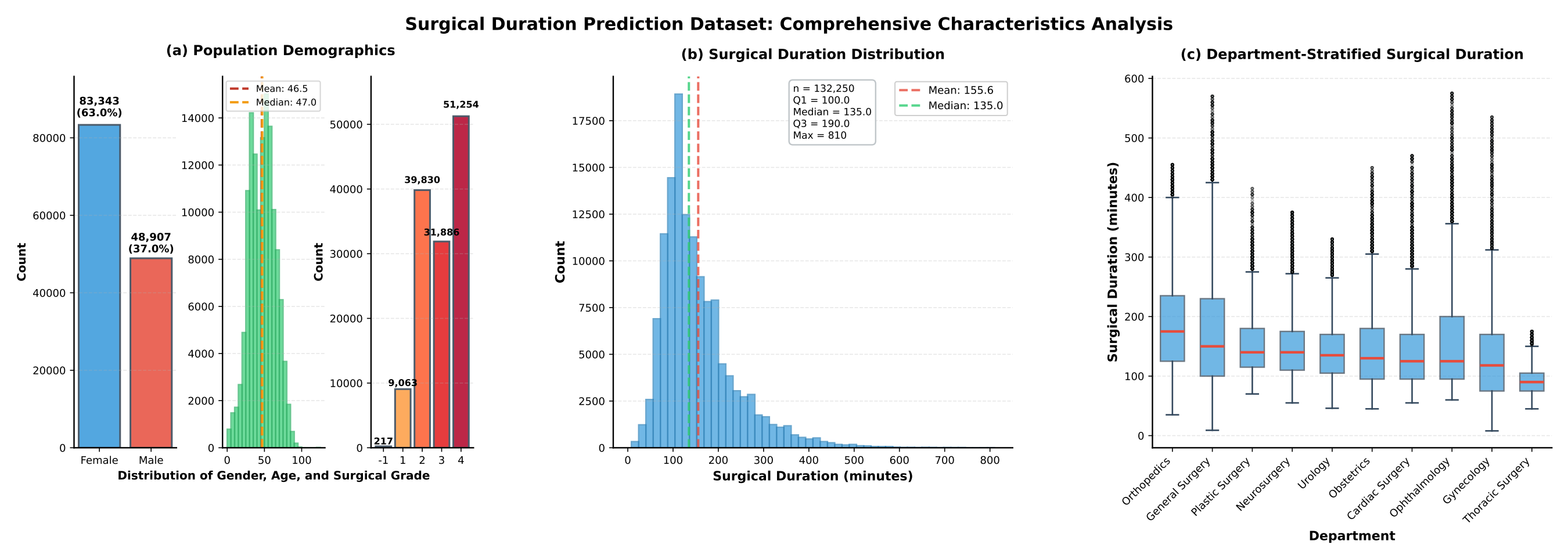}}
\caption{Key characteristics of In-hospital Dataset. (a) Patient and Surgical Characteristics: distributions of gender, age, and surgical grade. (b) Surgical duration distribution with key statistics. (c) Department-stratified duration patterns across 10 selected clinical departments.
}
\label{dataset_analysis}
\end{figure*}

\subsection{\textbf{Bayesian Averaging Aggregation}}\label{sec33} 
Multi-path LLM inference yields an ensemble of numeric candidates that reflect different plausible weighings of a case's clinical factors, but reconciling this dispersion into a single clinically credible estimate presents significant challenges. Classical aggregation rules, such as simple averaging and majority voting~\cite{vote_med}, stabilize random fluctuations but remain agnostic to institution-level knowledge, risking a consolidated estimate that departs from clinically plausible ranges.

To incorporate institution-specific knowledge, we introduce a Bayesian aggregation~\cite{bayesian} that combines the sampled LLM predictions with a population-level clinical prior. 
Those clinical precedent priors are derived from historical cases in the training set through hierarchical matching based on key clinical attributes.
Specifically, the matching process considers department, planned surgery name, and surgery level.
We compute the prior mean $\mu_{\text{prior}}$ as the median surgical duration, and adapt the prior strength $w_{\text{prior}}$ to the size of the matched cohort rather than fixing it.

This design grounds the estimate in the institution's own distribution rather than the sampled predictions alone, tempering estimates that would otherwise drift outside the clinically plausible range. 
Because the prior’s influence grows with the size of the matched cohort, well-represented procedures lean more on the institutional prior, while rarely seen procedures rely more on the sampled LLM outputs.

\section{Experiment}\label{sec5}
\subsection{\textbf{Experimental Setup}}\label{subsec51}

\subsubsection{Dataset}
To evaluate the effectiveness and generalizability of PREBA, we conduct experiments on four real-world datasets for surgical duration prediction. These datasets come from different healthcare institutions and span different patient populations and data distributions, allowing a thorough evaluation of PREBA across diverse settings.
\begin{itemize}

\item In-hospital dataset.
The in-hospital dataset was collected from the First People’s Hospital of Yunnan Province, China, covering surgical records from March 2022 to June 2025. After data preprocessing, the dataset contains 132,250 valid surgical cases from 51 clinical departments. As shown in Fig.~\ref{dataset_analysis} and Table~\ref{tab:dataset_summary}, the dataset includes 31 preoperative variables from multiple clinical dimensions, providing institution-specific surgical information for evaluating PREBA in real-world clinical scenarios.

\item MMSDP~\cite{xiuwen}.
The MMSDP dataset contains 88,538 surgical cases collected from a tertiary general hospital in China between June 2021 and August 2023. It provides 30 heterogeneous preoperative clinical features, including patient demographics, surgical characteristics, surgical teams, and laboratory examinations. It provides a representative benchmark for evaluating retrieval-based prediction using heterogeneous clinical information.

\item EPIC~\cite{mover}.
The EPIC dataset was collected from the University of California Irvine Medical Center through the Epic electronic health record system between 2017 and 2022. 
It contains 59,075 surgical cases and 6 categories of perioperative clinical variables, including patient demographics, surgical information, anesthesia-related variables, and laboratory measurements. 


\item INSPIRE~\cite{inspire}.
The INSPIRE dataset was collected from Seoul National University Hospital, South Korea, between 2011 and 2020. It contains 117K surgical cases with 10 categories of perioperative variables, including demographic information, surgical characteristics, anesthesia-related information, laboratory results, and clinical outcomes. 
The EPIC and INSPIRE datasets further enable evaluation of the cross-institutional generalization capability of PREBA.

\end{itemize}


The institutional dataset was collected under approved ethical protocols (Approval No. KHLL2023-KY032). All patient-identifiable information was removed or anonymized before model development. All experiments were conducted using de-identified clinical attributes and historical surgical cases, and no personally identifiable information was provided to the LLMs. The retrieved cases were used solely as contextual evidence for prediction generation.

\begin{table}[t]
\centering
\setlength{\tabcolsep}{5pt}
\caption{Surgical duration statistics across dataset splits}
\label{tab:dataset_summary}
\begin{tabular}{lccc}
\toprule
\textbf{Dataset} & \textbf{Count} & \textbf{Mean±Std} & \textbf{Range (min)} \\
\midrule
Train      & 125,369 & 155.8±83.7 & [8, 810] \\
Validation & 3,470   & 149.9±82.0 & [17, 720] \\
Test       & 3,411   & 154.3±81.2 & [11, 745] \\
\textbf{Total}       & 132,250   & 155.6±83.6 & [8, 810] \\
\bottomrule
\end{tabular}
\end{table}

\subsubsection{Baselines}\label{subsec:baselines}

To comprehensively evaluate PREBA, we compare it with representative approaches from four perspectives.

\begin{itemize}

\item LLM-based inference paradigms.
We evaluate three open-source LLMs, including Qwen3-8B~\cite{qwen3}, DeepSeek-R1-8B~\cite{deepseek}, and HuatuoGPT-o1-7B~\cite{huatuo-gpt}, under three inference protocols: \textit{Zero-Shot}, \textit{Random Few-Shot}, and \textit{Proposed RAG}. Zero-Shot directly predicts the target duration, while Random Few-Shot uses randomly selected historical cases as demonstrations. Proposed RAG incorporates PCA-weighted retrieved cases and statistical priors. 
We further evaluate Qwen3 with different parameter sizes under the proposed framework.

\item Machine learning.
We compare PREBA with nine fully-parametric supervised regression models, including Support Vector Regression (SVR)~\cite{svr}, Decision Tree~\cite{Decision_Tree}, Gradient Boosting Trees~\cite{Gradient_Boosting_Trees}, Bagging~\cite{Bagging}, Random Forest~\cite{randForest}, Quantile Random Forest~\cite{QrandForest}, Linear Regression~\cite{linear_regression}, Ridge Regression~\cite{Ridge_Regression}, and XGBoost~\cite{XGBoost}. 
All models use the same tabular feature for a fair comparison.

\item Alternative retrieval weighting strategies.
We compare PCA-based retrieval weighting with six alternative feature weighting strategies, including Uniform, Ridge coefficient~\cite{guyon2003introduction}, Mutual Information~\cite{walters2009estimation}, Permutation Importance~\cite{altmann2010permutation}, Random Forest Mean Decrease in Impurity (RF-MDI)~\cite{scornet2023trees}, and SHAP~\cite{mosca2022shap}. To isolate the effect of retrieval weighting, all strategies are evaluated using the same KNN-based mean aggregation.

\item Alternative aggregation strategies.
We compare Bayesian aggregation with various aggregation methods~\cite{self_consistency} under the same multi-path prediction setting. Mean, median, and majority voting aggregate sampled predictions through averaging, ranking, and frequency-based selection, respectively, while PREBA integrates multi-path predictions with clinical priors through Bayesian averaging.

\end{itemize}



\begin{table*}[t]
\centering
\caption{Main performance comparison across four datasets over five runs.}
\label{tab:main_results}
\begingroup
\footnotesize
\setlength{\tabcolsep}{1.3pt}
\renewcommand{\arraystretch}{1.0}
\setlength{\arrayrulewidth}{0.3pt}
\setlength{\aboverulesep}{1.8pt}
\setlength{\belowrulesep}{2.2pt}
\newcommand{\prebaresult}[2]{\shortstack[c]{\strut#1\\[-3pt]{\normalfont\fontsize{6}{6.5}\selectfont\color{black!60}$\pm$\,#2}}}
\begin{tabular*}{\textwidth}{@{\extracolsep{\fill}}ll|*{4}{c}@{\hspace{7pt}}*{4}{c}@{\hspace{7pt}}*{4}{c}@{\hspace{7pt}}*{4}{c}@{}}
\toprule
\multirow{2}{*}{Setting} & \multirow{2}{*}{Model} & \multicolumn{4}{c}{\textbf{In-hospital Dataset}} & \multicolumn{4}{c}{\textbf{MMSDP~\cite{xiuwen}}} & \multicolumn{4}{c}{\textbf{INSPIRE~\cite{inspire}}} & \multicolumn{4}{c}{\textbf{EPIC~\cite{mover}}} \\
\cmidrule(lr){3-6}\cmidrule(lr){7-10}\cmidrule(lr){11-14}\cmidrule(l){15-18}
 & & MAE & RMSE & $R^2$ & MAPE & MAE & RMSE & $R^2$ & MAPE & MAE & RMSE & $R^2$ & MAPE & MAE & RMSE & $R^2$ & MAPE \\
\midrule
 & Qwen3-8B~\cite{qwen3} & \prebaresult{59.87}{1.057} & \prebaresult{88.22}{1.482} & \prebaresult{-0.18}{0.167} & \prebaresult{38.81}{1.030} & \prebaresult{79.35}{0.333} & \prebaresult{94.08}{0.388} & \prebaresult{-2.17}{0.086} & \prebaresult{41.68}{0.249} & \prebaresult{66.29}{0.583} & \prebaresult{91.63}{0.156} & \prebaresult{0.23}{0.003} & \prebaresult{122.69}{3.757} & \prebaresult{95.13}{0.245} & \prebaresult{151.20}{0.227} & \prebaresult{-0.05}{0.003} & \prebaresult{\underline{40.01}}{0.122} \\[2pt]
Zero-Shot & DeepSeek-R1-8B~\cite{deepseek} & \prebaresult{64.05}{1.109} & \prebaresult{96.07}{2.040} & \prebaresult{-0.40}{0.051} & \prebaresult{39.11}{0.376} & \prebaresult{84.47}{0.234} & \prebaresult{109.18}{0.174} & \prebaresult{-3.27}{0.198} & \prebaresult{44.93}{0.229} & \prebaresult{69.70}{0.339} & \prebaresult{98.61}{0.192} & \prebaresult{0.11}{0.005} & \prebaresult{121.42}{2.733} & \prebaresult{104.56}{0.943} & \prebaresult{161.08}{0.623} & \prebaresult{-0.19}{0.009} & \prebaresult{47.79}{0.716} \\[2pt]
 & HuatuoGPT-o1-7B~\cite{huatuo-gpt} & \prebaresult{52.63}{0.499} & \prebaresult{76.65}{0.465} & \prebaresult{0.11}{0.017} & \prebaresult{32.77}{0.131} & \prebaresult{82.03}{0.396} & \prebaresult{102.10}{1.332} & \prebaresult{-2.74}{0.435} & \prebaresult{42.40}{0.336} & \prebaresult{79.35}{0.302} & \prebaresult{102.86}{0.163} & \prebaresult{0.03}{0.004} & \prebaresult{173.53}{1.216} & \prebaresult{100.06}{0.070} & \prebaresult{157.56}{0.072} & \prebaresult{-0.14}{0.001} & \prebaresult{49.01}{0.102} \\[2pt]
\midrule[0.3pt]
 & Qwen3-8B~\cite{qwen3} & \prebaresult{48.96}{0.197} & \prebaresult{74.97}{0.229} & \prebaresult{0.14}{0.013} & \prebaresult{31.40}{0.011} & \prebaresult{52.68}{0.171} & \prebaresult{73.22}{0.198} & \prebaresult{-0.92}{0.007} & \prebaresult{29.46}{0.067} & \prebaresult{66.12}{0.178} & \prebaresult{98.46}{0.400} & \prebaresult{0.12}{0.007} & \prebaresult{106.83}{2.816} & \prebaresult{86.09}{0.272} & \prebaresult{135.89}{0.450} & \prebaresult{0.15}{0.006} & \prebaresult{46.69}{0.163} \\[2pt]
\shortstack[l]{Random\\Few-Shot} & DeepSeek-R1-8B~\cite{deepseek} & \prebaresult{54.30}{0.137} & \prebaresult{81.71}{0.264} & \prebaresult{-0.01}{0.005} & \prebaresult{32.22}{0.016} & \prebaresult{48.29}{0.263} & \prebaresult{65.89}{0.418} & \prebaresult{-0.56}{0.010} & \prebaresult{25.69}{0.228} & \prebaresult{66.09}{0.175} & \prebaresult{99.69}{0.525} & \prebaresult{0.09}{0.009} & \prebaresult{97.75}{2.987} & \prebaresult{98.01}{0.447} & \prebaresult{141.65}{0.203} & \prebaresult{0.08}{0.003} & \prebaresult{69.01}{0.869} \\[2pt]
 & HuatuoGPT-o1-7B~\cite{huatuo-gpt} & \prebaresult{47.22}{0.235} & \prebaresult{67.96}{0.220} & \prebaresult{0.29}{0.015} & \prebaresult{31.32}{0.046} & \prebaresult{55.97}{0.146} & \prebaresult{72.77}{0.308} & \prebaresult{-0.90}{0.015} & \prebaresult{30.02}{0.158} & \prebaresult{77.60}{0.122} & \prebaresult{102.64}{0.167} & \prebaresult{0.04}{0.003} & \prebaresult{164.80}{1.088} & \prebaresult{94.05}{0.190} & \prebaresult{143.06}{0.086} & \prebaresult{0.06}{0.001} & \prebaresult{54.51}{0.209} \\[2pt]
\midrule[0.3pt]
 & Qwen3-8B~\cite{qwen3} & \prebaresult{\underline{35.81}}{0.175} & \prebaresult{\underline{50.11}}{0.303} & \prebaresult{\underline{0.62}}{0.006} & \prebaresult{26.13}{0.066} & \prebaresult{\underline{28.48}}{0.078} & \prebaresult{\underline{42.77}}{0.143} & \prebaresult{\underline{0.34}}{0.004} & \prebaresult{18.03}{0.105} & \prebaresult{\underline{40.60}}{0.074} & \prebaresult{\textbf{65.78}}{0.296} & \prebaresult{\textbf{0.61}}{0.004} & \prebaresult{\underline{50.51}}{0.346} & \prebaresult{\textbf{68.07}}{0.293} & \prebaresult{\textbf{112.37}}{0.283} & \prebaresult{\textbf{0.42}}{0.003} & \prebaresult{41.81}{0.422} \\[2pt]
\shortstack[l]{PREBA\\(Ours)} & DeepSeek-R1-8B~\cite{deepseek} & \prebaresult{35.91}{0.150} & \prebaresult{50.90}{0.201} & \prebaresult{0.61}{0.003} & \prebaresult{\textbf{25.74}}{0.076} & \prebaresult{29.16}{0.042} & \prebaresult{44.66}{0.112} & \prebaresult{0.28}{0.003} & \prebaresult{\underline{17.97}}{0.068} & \prebaresult{\textbf{40.25}}{0.450} & \prebaresult{\underline{67.72}}{0.494} & \prebaresult{\underline{0.58}}{0.007} & \prebaresult{\textbf{47.51}}{3.985} & \prebaresult{70.11}{0.130} & \prebaresult{\underline{118.10}}{0.484} & \prebaresult{\underline{0.36}}{0.005} & \prebaresult{40.63}{0.353} \\[2pt]
 & HuatuoGPT-o1-7B~\cite{huatuo-gpt} & \prebaresult{\textbf{35.52}}{0.141} & \prebaresult{\textbf{49.55}}{0.222} & \prebaresult{\textbf{0.63}}{0.002} & \prebaresult{\underline{25.78}}{0.064} & \prebaresult{\textbf{27.85}}{0.049} & \prebaresult{\textbf{41.33}}{0.184} & \prebaresult{\textbf{0.39}}{0.004} & \prebaresult{\textbf{17.32}}{0.020} & \prebaresult{43.11}{0.184} & \prebaresult{69.83}{0.827} & \prebaresult{0.55}{0.011} & \prebaresult{56.26}{1.052} & \prebaresult{\underline{70.04}}{0.171} & \prebaresult{118.77}{1.067} & \prebaresult{0.35}{0.012} & \prebaresult{\textbf{39.50}}{0.485} \\[2pt]
\bottomrule
\end{tabular*}
\par\vspace{3pt}
\noindent
\begin{minipage}{\textwidth}
\fontsize{7}{8}\selectfont
\textit{Note:} Each cell reports the mean (top) and $\pm$ standard deviation (bottom) over five independent runs. MAE and RMSE are in minutes; MAPE is in percent. Lower MAE, RMSE, and MAPE and higher $R^2$ indicate better performance. The best and second-best means in each column are \textbf{bold} and \underline{underlined}, respectively.
\end{minipage}
\endgroup
\end{table*}

\subsubsection{Evaluation Metrics}
\label{subsubsec512}

To comprehensively evaluate model performance, we employ four established regression metrics: Mean Absolute Error (MAE), Root Mean Square Error (RMSE), Coefficient of Determination ($R^2$), and Mean Absolute Percentage Error (MAPE). These metrics provide complementary insights into different aspects of prediction accuracy.

Mean Absolute Error quantifies the average magnitude of prediction errors in absolute units:
\begin{equation}
\text{MAE} = \frac{1}{m}\sum_{i=1}^{m}|y_i - \hat{y}_i|
\end{equation}
where $m$ is the number of test samples, $y_i$ is the true surgical duration, and $\hat{y}_i$ is the predicted duration. 

Root Mean Square Error penalizes larger errors more heavily than MAE:
\begin{equation}
\text{RMSE} = \sqrt{\frac{1}{m}\sum_{i=1}^{m}(y_i - \hat{y}_i)^2}
\end{equation}
RMSE is sensitive to outlier predictions and reveals whether the model exhibits occasional severe errors. A high RMSE relative to MAE signals the presence of large prediction failures on specific cases, which is critical to identify in clinical contexts where consistency matters.

The coefficient of determination measures the proportion of variance in surgical durations explained by the model:
\begin{equation}
R^2 = 1 - \frac{\sum_{i=1}^{m}(y_i - \hat{y}_i)^2}{\sum_{i=1}^{m}(y_i - \bar{y})^2}
\end{equation}
where $\bar{y}$ is the mean of observed durations. $R^2$ quantifies overall model fit; values closer to 1 indicate that the model captures the underlying patterns in duration variation.

Mean Absolute Percentage Error expresses prediction error as a percentage of true values:
\begin{equation}
\text{MAPE} = \frac{1}{m}\sum_{i=1}^{m}\left|\frac{y_i - \hat{y}_i}{y_i}\right| \times 100\%
\end{equation}

\begin{table*}[t]
\centering
\caption{Performance comparison with traditional machine learning methods on MMSDP and EPIC over five runs.}
\label{tab:ml_results}
\begingroup
\footnotesize
\setlength{\tabcolsep}{1.3pt}
\renewcommand{\arraystretch}{1.12}
\setlength{\aboverulesep}{1.8pt}
\setlength{\belowrulesep}{2.2pt}
\newcommand{\mlresult}[2]{\mbox{#1\,{\normalfont\fontsize{6}{7}\selectfont\color{black!60}$\pm$\,#2}}}
\begin{tabular*}{\textwidth}{@{\extracolsep{\fill}}ll|*{4}{r}@{\hspace{7pt}}*{4}{r}@{}}
\toprule
\multirow{2}{*}{Setting} & \multirow{2}{*}{Model}
 & \multicolumn{4}{c}{\textbf{MMSDP}~\cite{xiuwen}}
 & \multicolumn{4}{c}{\textbf{EPIC}~\cite{mover}} \\
\cmidrule(lr){3-6}\cmidrule(l){7-10}
 & & MAE $\downarrow$ & RMSE $\downarrow$ & $R^2$ $\uparrow$ & MAPE $\downarrow$
   & MAE $\downarrow$ & RMSE $\downarrow$ & $R^2$ $\uparrow$ & MAPE $\downarrow$ \\
\midrule
\multirow{9}{*}{\shortstack[l]{Full-Parameter \\ Training ML}} & Linear Regression~\cite{linear_regression}
 & \mlresult{42.63}{0.963} & \mlresult{57.74}{0.679} & \mlresult{-0.17}{0.041} & \mlresult{23.63}{0.388} & \mlresult{81.97}{0.570} & \mlresult{121.72}{1.083} & \mlresult{0.32}{0.030} & \mlresult{55.30}{0.434} \\
 & Ridge Regression~\cite{Ridge_Regression}
 & \mlresult{39.96}{0.597} & \mlresult{55.50}{0.358} & \mlresult{-0.10}{0.014} & \mlresult{21.36}{0.239} & \mlresult{81.83}{0.938} & \mlresult{122.47}{1.864} & \mlresult{0.32}{0.031} & \mlresult{55.16}{1.080} \\
 & SVR~\cite{svr}
 & \mlresult{34.44}{0.962} & \mlresult{50.93}{0.729} & \mlresult{0.08}{0.032} & \mlresult{18.47}{0.351} & \mlresult{88.46}{0.701} & \mlresult{141.19}{0.768} & \mlresult{0.08}{0.010} & \mlresult{51.19}{0.427} \\
 & Decision Tree~\cite{Decision_Tree}
 & \mlresult{49.38}{0.482} & \mlresult{70.34}{0.567} & \mlresult{-0.78}{0.114} & \mlresult{27.82}{0.318} & \mlresult{116.77}{1.381} & \mlresult{184.51}{2.390} & \mlresult{-0.56}{0.133} & \mlresult{76.80}{1.124} \\
 & Bagging~\cite{Bagging}
 & \mlresult{35.23}{0.709} & \mlresult{51.70}{0.428} & \mlresult{0.05}{0.023} & \mlresult{18.75}{0.189} & \mlresult{83.65}{1.202} & \mlresult{125.93}{1.470} & \mlresult{0.27}{0.037} & \mlresult{59.39}{0.493} \\
 & Random Forest~\cite{randForest}
 & \mlresult{34.51}{0.696} & \mlresult{49.75}{0.569} & \mlresult{0.11}{0.020} & \mlresult{18.85}{0.312} & \mlresult{85.11}{0.768} & \mlresult{127.08}{1.128} & \mlresult{0.26}{0.019} & \mlresult{62.29}{0.172} \\
 & Quantile Random Forest~\cite{QrandForest}
 & \mlresult{34.21}{0.716} & \mlresult{51.07}{0.435} & \mlresult{0.09}{0.015} & \mlresult{18.19}{0.113} & \mlresult{85.60}{1.210} & \mlresult{127.68}{1.707} & \mlresult{0.25}{0.027} & \mlresult{62.37}{0.357} \\
 & Gradient Boosting Trees~\cite{Gradient_Boosting_Trees}
 & \mlresult{32.29}{0.806} & \mlresult{46.32}{0.833} & \mlresult{0.24}{0.022} & \mlresult{18.54}{0.278} & \mlresult{82.71}{1.106} & \mlresult{123.99}{2.092} & \mlresult{0.30}{0.027} & \mlresult{56.95}{0.631} \\
 & XGBoost~\cite{XGBoost}
 & \mlresult{31.20}{0.565} & \mlresult{46.42}{0.500} & \mlresult{0.26}{0.012} & \mlresult{\textbf{16.85}}{0.074} & \mlresult{75.12}{0.828} & \mlresult{118.24}{1.116} & \mlresult{\underline{0.36}}{0.023} & \mlresult{49.70}{0.335} \\
\midrule[0.3pt]
\multirow{3}{*}{\shortstack[l]{PREBA\\(Ours)}} & Qwen3-8B~\cite{qwen3}
 & \mlresult{\underline{28.48}}{0.078} & \mlresult{\underline{42.77}}{0.143} & \mlresult{\underline{0.34}}{0.004} & \mlresult{18.03}{0.105} & \mlresult{\textbf{68.07}}{0.293} & \mlresult{\textbf{112.37}}{0.283} & \mlresult{\textbf{0.42}}{0.003} & \mlresult{41.81}{0.422} \\
 & DeepSeek-R1-8B~\cite{deepseek}
 & \mlresult{29.16}{0.042} & \mlresult{44.66}{0.112} & \mlresult{0.28}{0.003} & \mlresult{17.97}{0.068} & \mlresult{70.11}{0.130} & \mlresult{\underline{118.10}}{0.484} & \mlresult{\underline{0.36}}{0.005} & \mlresult{\underline{40.63}}{0.353} \\
 & HuatuoGPT-o1-7B~\cite{huatuo-gpt}
 & \mlresult{\textbf{27.85}}{0.049} & \mlresult{\textbf{41.33}}{0.184} & \mlresult{\textbf{0.39}}{0.004} & \mlresult{\underline{17.32}}{0.020} & \mlresult{\underline{70.04}}{0.171} & \mlresult{118.77}{1.067} & \mlresult{0.35}{0.012} & \mlresult{\textbf{39.50}}{0.485} \\
\bottomrule
\end{tabular*}
\par\vspace{3pt}
\noindent
\begin{minipage}{\textwidth}
\fontsize{7}{8.5}\selectfont
\textit{Note:} Values are mean $\pm$ standard deviation over five independent runs.
MAE and RMSE are in minutes; MAPE is in percent.
$\downarrow$/$\uparrow$ indicate lower/higher is better.
The best and second-best means in each column are \textbf{bold} and
\underline{underlined}, respectively; ties at the reported precision share the same formatting.
\end{minipage}
\endgroup
\end{table*}

MAE and MAPE evaluate model performance along complementary dimensions. MAE quantifies the absolute prediction error in minutes. MAPE, in contrast, assesses relative accuracy across the full spectrum of surgical complexity. 

\subsubsection{Implementation Details}
\label{subsec:implementation}
For the proposed RAG setting, the retrieval database is constructed using only training cases to prevent information leakage. The top-$K$ retrieval size is set to 8. To improve retrieval robustness, we first retrieve 80 candidate cases (10$\times$ expansion) using FAISS with the IndexFlat configuration~\cite{faiss}, and then select the final 8 references through clinical filtering.
All LLM-based experiments employ multi-path prediction with $n=5$ sampling rounds for each query. Each sampling round independently generates one prediction using the same prompt template and input format. The first round adopts deterministic decoding ($\tau_1=0$), while the remaining rounds use randomly sampled temperatures from $\mathcal{U}(0.05,0.4)$. 
For PREBA, Bayesian aggregation adaptively integrates multi-path predictions with retrieved clinical evidence and prior information. 
All ML baselines are trained on the raw structured tabular features of each dataset, with numerical features used directly and categorical, ordinal and textual fields integer-encoded, which is the representation best suited to tree-based models. 
During inference, 10 parallel threads are used to accelerate prediction. 
Experiments are conducted on a server equipped with four NVIDIA A100 40GB GPUs, with LLM inference accelerated by vLLM and LlamaFactory tool.

\begin{table}[t]
\centering
\caption{Comparison of retrieval weighting methods.}
\label{tab:retrieval_weighting}
\begingroup
\fontsize{7.5}{9}\selectfont
\setlength{\tabcolsep}{0.6pt}
\setlength{\arrayrulewidth}{0.3pt}
\renewcommand{\arraystretch}{1.15}
\setlength{\aboverulesep}{1.5pt}
\setlength{\belowrulesep}{2pt}
\begin{tabular*}{\columnwidth}{@{\extracolsep{\fill}}cl*{4}{c}|*{4}{c}@{}}
\toprule
\multirow{2}{*}{Label} & \multirow{2}{*}{Weighting}
 & \multicolumn{4}{c|}{\textbf{MMSDP}~\cite{xiuwen}}
 & \multicolumn{4}{c}{\textbf{EPIC}~\cite{mover}} \\
\cmidrule(lr){3-6}\cmidrule(l){7-10}
 & & MAE & RMSE & $R^2$ & MAPE & MAE & RMSE & $R^2$ & MAPE \\
\midrule
\multirow{5}{*}{Yes} & Ridge coef. & 30.07 & 45.82 & 0.24 & 18.49 & 71.05 & 117.48 & \underline{0.36} & 41.52 \\
 & Mutual info. & \underline{29.88} & \underline{45.81} & \underline{0.26} & \textbf{17.89} & \underline{70.91} & 120.73 & 0.33 & 41.49 \\
 & Permutation & 32.39 & 48.94 & 0.14 & 19.47 & \textbf{68.27} & \textbf{113.19} & \textbf{0.41} & \textbf{37.31} \\
 & RF-MDI & 32.50 & 48.77 & 0.15 & 20.02 & 72.87 & 121.41 & 0.33 & 42.17 \\
 & SHAP & 32.34 & 48.55 & 0.15 & 19.88 & 74.25 & 122.76 & 0.31 & 43.96 \\
\midrule[0.3pt]
\multirow{2}{*}{No} & Uniform & 35.35 & 52.22 & 0.05 & 22.10 & 78.93 & 127.34 & 0.25 & 44.78 \\
 & PCA (Ours) & \textbf{29.57} & \textbf{45.27} & \textbf{0.27} & \underline{18.26} & 71.02 & \underline{117.45} & \underline{0.36} & \underline{38.99} \\
\bottomrule
\end{tabular*}
\par\vspace{3pt}
\noindent
\begin{minipage}{\columnwidth}
\fontsize{7}{8.5}\selectfont
\textit{Note:} Label indicates use of duration labels to derive retrieval weights.
RF-MDI: random-forest mean decrease in impurity.
\end{minipage}
\endgroup
\end{table}

\subsection{\textbf{Performance Comparison}}\label{subsec53}

PREBA consistently outperforms zero-shot and random few-shot inference paradigms across four real-world datasets. As shown in Table~\ref{tab:main_results}, zero-shot setting generally yields the weakest performance, reflecting the limited ability of LLMs to estimate surgical duration without institution-specific evidence. Random few-shot setting improves prediction accuracy by incorporating historical cases. 
On In-hospital Dataset, for example, the MAE of Qwen3-8B decreases from $59.87$ to $48.96$ minutes.
PREBA further reduces MAE to $35.81$ minutes, representing relative reductions of $40.2\%$ and $26.9\%$ over zero-shot and random few-shot inference, respectively. 
Similar trends are observed on MMSDP, INSPIRE, and EPIC, where PREBA reduces MAE by up to $64.9\%$, $39.3\%$, and $28.4\%$, respectively, relative to the strongest zero-shot configuration. These consistent gains reflect the combined effect of grounding predictions in similar institution-specific evidence and aggregating multi-path LLM outputs with clinical statistical priors, which together yield calibrated and stable duration estimates.

PREBA achieves competitive or superior performance relative to the fully supervised baselines on MMSDP and EPIC.
As shown in Table~\ref{tab:ml_results}, on MMSDP, HuatuoGPT-o1-7B with PREBA achieves an MAE of $27.85\pm0.05$ minutes, RMSE of $41.33\pm0.18$ minutes, and $R^2$ of $0.39\pm0.004$, surpassing XGBoost, the strongest ML baseline, which achieves an MAE of $31.20\pm0.57$ minutes, RMSE of $46.42\pm0.50$ minutes, and $R^2$ of $0.26\pm0.012$. 
On EPIC, Qwen3-8B with PREBA attains an MAE of $68.07\pm0.29$ minutes and an $R^2$ of $0.42\pm0.003$, improving upon XGBoost by $9.4\%$ in MAE and increasing $R^2$ from $0.36$ to $0.42$. 
These results show that PREBA can achieve superior prediction performance without task-specific LLM fine-tuning. 
Importantly, it provides explicit clinical interpretability through retrieved cases and statistical priors, and enables seamless adaptation to new institutional contexts. 

\subsection{\textbf{Comparison with Alternative Retrieval Weighting and Aggregation Strategies}}
\label{subsec:alternative_strategies}
PCA weighting provides an effective retrieval weighting strategy without requiring duration labels for weight estimation. As shown in Table~\ref{tab:retrieval_weighting}, on MMSDP, PCA reduces MAE from $35.35$ minutes with uniform weighting to $29.57$ minutes and improves $R^2$ from $0.05$ to $0.27$, achieving the lowest MAE and highest $R^2$ among all compared strategies. On EPIC, PCA reduces MAE by $10.0\%$ relative to uniform weighting ($78.93$ to $71.02$ minutes) and achieves an $R^2$ of $0.36$, matching Ridge coefficient weighting and remaining competitive with the supervised alternatives. 
While label-dependent and model-based alternatives require outcome supervision and, for some methods, an additional fitted predictor or repeated importance evaluation, PCA estimates retrieval weights directly from historical clinical representations without using duration labels. 
These results support PCA weighting as an effective retrieval strategy for PREBA, offering performance comparable to supervised alternatives while avoiding label-dependent weight estimation.

Bayesian averaging consistently achieves the lowest MAE among the compared aggregation strategies. As shown in Table~\ref{tab:aggregation}, it improves upon simple averaging for both DeepSeek-R1-8B and HuatuoGPT-o1-7B on MMSDP and EPIC. Specifically, Bayesian averaging reduces the MAE of DeepSeek-R1-8B from $30.78$ to $29.16$ minutes on MMSDP and from $71.57$ to $70.11$ minutes on EPIC. For HuatuoGPT-o1-7B, it similarly reduces the MAE from $29.60$ to $27.85$ minutes on MMSDP and from $71.97$ to $70.04$ minutes on EPIC. Unlike compared aggregation rules based solely on generated outputs, Bayesian averaging incorporates clinical statistical priors to consolidate multi-path predictions, reducing the influence of sampling variation and yielding more calibrated and stable duration estimates.

\begin{table}[t]
\centering
\caption{Comparison of prediction aggregation strategies.}
\label{tab:aggregation}
\begingroup
\fontsize{7}{8.5}\selectfont
\setlength{\tabcolsep}{0.6pt}
\setlength{\arrayrulewidth}{0.3pt}
\renewcommand{\arraystretch}{1.15}
\setlength{\aboverulesep}{1.5pt}
\setlength{\belowrulesep}{2pt}
\begin{tabular*}{\columnwidth}{@{\extracolsep{\fill}}ll*{4}{c}|*{4}{c}@{}}
\toprule
\multirow{2}{*}{Model} & \multirow{2}{*}{Strategy}
 & \multicolumn{4}{c|}{\textbf{MMSDP}~\cite{xiuwen}}
 & \multicolumn{4}{c}{\textbf{EPIC}~\cite{mover}} \\
\cmidrule(lr){3-6}\cmidrule(l){7-10}
 & & MAE & RMSE & $R^2$ & MAPE
 & MAE & RMSE & $R^2$ & MAPE \\
\midrule
\multirow{5}{*}{\shortstack[l]{DeepSeek-\\R1-8B~\cite{deepseek}}} & Median & \underline{30.55} & 45.87 & \underline{0.27} & 18.55 & 71.84 & 120.51 & \underline{0.33} & 41.36 \\
 & Majority vote & 31.22 & 46.84 & 0.23 & 18.71 & 72.23 & 121.66 & 0.30 & 41.70 \\
 & Quantile avg. & 30.71 & \underline{45.77} & 0.25 & \underline{18.52} & \underline{71.57} & \underline{120.12} & 0.32 & 41.43 \\
 & Simple avg. & 30.78 & 45.95 & \underline{0.27} & 18.57 & \underline{71.57} & 120.65 & 0.32 & \textbf{40.26} \\
 & Bayesian (Ours) & \textbf{29.16} & \textbf{44.66} & \textbf{0.28} & \textbf{17.97} & \textbf{70.11} & \textbf{118.10} & \textbf{0.36} & \underline{40.63} \\
\midrule[0.5pt]
\multirow{5}{*}{\shortstack[l]{HuatuoGPT-\\o1-7B~\cite{huatuo-gpt}}} & Median & \underline{29.56} & 42.82 & 0.35 & 18.42 & 71.86 & 123.14 & 0.30 & 40.79 \\
 & Majority vote & 29.90 & 43.43 & 0.33 & 18.60 & 72.46 & 125.49 & 0.27 & \underline{40.63} \\
 & Quantile avg. & 29.90 & \underline{42.58} & \underline{0.36} & \underline{18.34} & \underline{71.21} & \underline{121.49} & \underline{0.31} & 40.82 \\
 & Simple avg. & 29.60 & 42.92 & 0.35 & 18.44 & 71.97 & 123.51 & 0.30 & 40.65 \\
 & Bayesian (Ours) & \textbf{27.85} & \textbf{41.33} & \textbf{0.39} & \textbf{17.32} & \textbf{70.04} & \textbf{118.77} & \textbf{0.35} & \textbf{39.50} \\
\bottomrule
\end{tabular*}
\endgroup
\end{table}

\begin{figure*}[t]
\centering
\centerline{\includegraphics[width=1\linewidth]{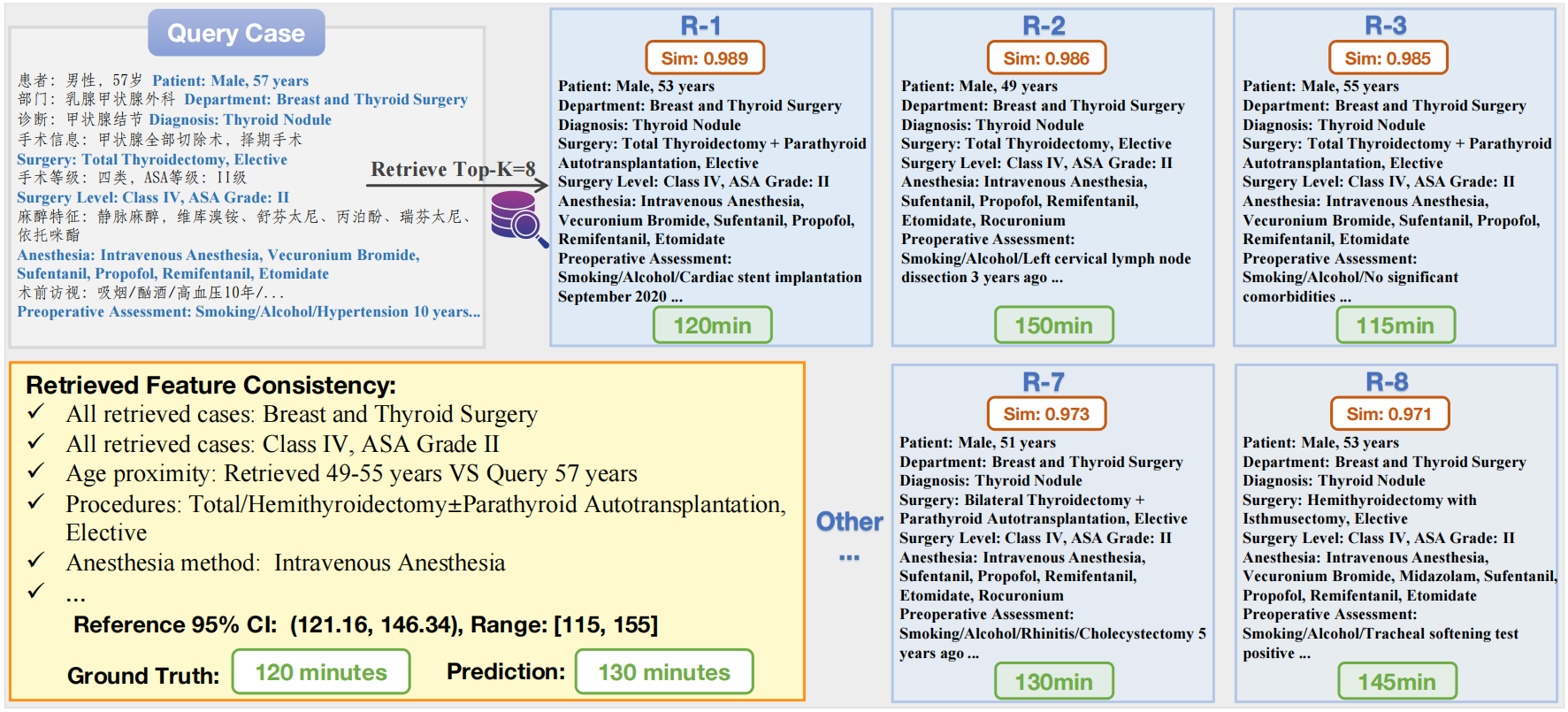}}
\caption{Case study visualizations of PREBA on the in-hospital dataset using Qwen3-8B. The figure shows one query case (grey) and its retrieved  surgeries (blue), together with their similarity scores and observed durations, illustrating how institution-specific evidence is assembled for the LLM. The bottom panel (yellow) summarizes feature consistency and reference 95\% CI and ground-truth duration with the PREBA prediction.}
\label{case_study}
\end{figure*}

\begin{figure}[t]
\centering
\centerline{\includegraphics[width=1\linewidth]{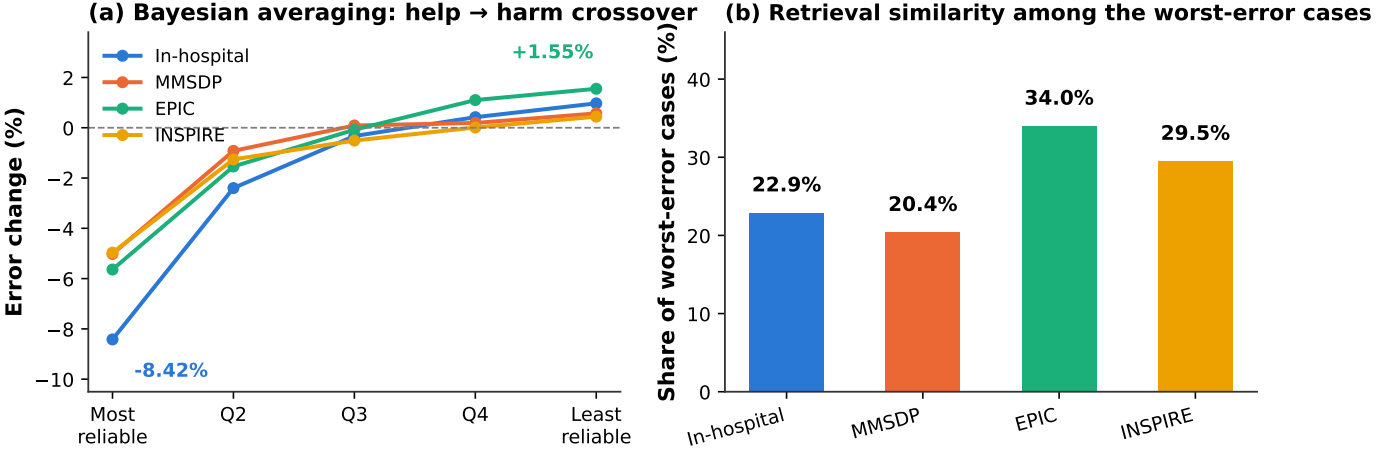}}
\caption{Analysis of PREBA. (a) 
Bayesian averaging reduces error when retrieved evidence is reliable, 
but increases it when evidence deviates most. (b) Among the 5\% of 
cases with the largest errors, 20.4\%-34.0\% retrieve cases that look 
similar but are not the query's true match.}
\label{fig:failure}
\end{figure}

\begin{table}[t]
\centering
\caption{Ablation study on the effect of Top-$k$ retrieval.}
\label{tab:topk_ablation}
\begingroup
\fontsize{7}{8.5}\selectfont
\setlength{\tabcolsep}{0pt}
\setlength{\arrayrulewidth}{0.3pt}
\renewcommand{\arraystretch}{1.15}
\setlength{\aboverulesep}{1.5pt}
\setlength{\belowrulesep}{2pt}

\newcommand{\topkmetricwidth}{\dimexpr(\columnwidth-53pt-\arrayrulewidth)/10\relax}

\begin{tabular}{@{}>{\raggedright\arraybackslash}m{44pt}>{\centering\arraybackslash}m{9pt}*{5}{>{\centering\arraybackslash}m{\topkmetricwidth}}|*{5}{>{\centering\arraybackslash}m{\topkmetricwidth}}@{}}
\toprule
\multirow{2}{*}{Model} & \multirow{2}{*}{$k$}
& \multicolumn{5}{c|}{\textbf{MMSDP}~\cite{xiuwen}}
& \multicolumn{5}{c}{\textbf{EPIC}~\cite{mover}} \\
\cmidrule(lr){3-7}\cmidrule(l){8-12}
& & MAE & RMSE & $R^2$ & MAPE & Time
& MAE & RMSE & $R^2$ & MAPE & Time \\
\midrule

& 2
& 30.23 & 45.07 & 0.27 & 19.07 & 3.59
& 72.77 & 122.21 & 0.31 & 43.14 & \textbf{2.58} \\

& 4
& \underline{29.39} & \underline{44.34} & \textbf{0.30} & 18.32 & \underline{3.86}
& 70.71 & 119.20 & \underline{0.35} & 41.36 & 2.69 \\

& 6
& 29.47 & 44.70 & 0.28 & 18.30 & 4.92
& \underline{70.15} & 119.44 & 0.34 & \underline{40.38} & 3.03 \\

\rowcolor{black!8}
\cellcolor{white} & 8
& \textbf{29.16} & \textbf{44.66} & 0.28 & \textbf{17.97} & \textbf{3.02}
& \underline{70.11} & \underline{118.10} & \underline{0.36} & 40.63 & \underline{2.74} \\

\multirow{-5}{*}{\shortstack[l]{DeepSeek-\\R1-8B~\cite{deepseek}}}
& 10
& 29.40 & 44.90 & 0.28 & \underline{18.09} & 5.62
& \textbf{69.23} & \textbf{116.99} & \textbf{0.37} & \textbf{39.85} & 3.20 \\

\midrule[0.3pt]

& 2
& 28.81 & 43.02 & 0.34 & 17.55 & \textbf{2.67}
& 73.11 & 124.14 & 0.29 & 41.53 & \textbf{1.89} \\

& 4
& 28.11 & 42.10 & 0.36 & \underline{17.23} & \underline{3.61}
& 72.38 & 122.15 & 0.31 & 41.31 & \underline{1.92} \\

& 6
& 27.80 & 41.44 & 0.38 & \textbf{17.19} & 3.88
& 71.35 & 121.61 & 0.32 & \underline{40.32} & 2.23 \\

\rowcolor{black!8}
\cellcolor{white} & 8
& \textbf{27.85} & \underline{41.33} & \textbf{0.39} & 17.32 & 4.69
& \textbf{70.04} & \textbf{118.77} & \textbf{0.35} & \textbf{39.50} & 2.63 \\

\multirow{-5}{*}{\shortstack[l]{HuatuoGPT-\\o1-7B~\cite{huatuo-gpt}}}
& 10
& \underline{27.86} & \textbf{41.27} & \textbf{0.39} & 17.44 & \underline{4.60}
& \underline{70.06} & \underline{119.58} & \underline{0.34} & 40.35 & 2.66 \\

\bottomrule
\end{tabular}
\endgroup
\end{table}

\begin{table}[t]
\centering
\caption{Ablation study of PCA-weighted retrieval.}
\label{tab:pca_ablation}
\begingroup
\fontsize{7.5}{9}\selectfont
\setlength{\tabcolsep}{0.6pt}
\setlength{\arrayrulewidth}{0.3pt}
\renewcommand{\arraystretch}{1.2}
\setlength{\aboverulesep}{1.5pt}
\setlength{\belowrulesep}{2pt}
\begin{tabular*}{\columnwidth}{@{\extracolsep{\fill}}lc*{4}{c}|*{4}{c}@{}}
\toprule
\multirow{2}{*}{Model} & \multirow{2}{*}{PCA}
 & \multicolumn{4}{c|}{\textbf{MMSDP}~\cite{xiuwen}}
 & \multicolumn{4}{c}{\textbf{EPIC}~\cite{mover}} \\
\cmidrule(lr){3-6}\cmidrule(l){7-10}
 & & MAE & RMSE & $R^2$ & MAPE & MAE & RMSE & $R^2$ & MAPE \\
\midrule
\multirow{2}{*}{\shortstack[l]{DeepSeek-\\R1-8B~\cite{deepseek}}} & w/o & 31.79 & 47.71 & 0.20 & 18.88 & 73.68 & 126.18 & 0.25 & 41.35 \\
 & w/ & \textbf{29.16} & \textbf{44.66} & \textbf{0.28} & \textbf{17.97} & \textbf{70.11} & \textbf{118.10} & \textbf{0.36} & \textbf{40.63} \\
\midrule[0.3pt]
\multirow{2}{*}{\shortstack[l]{HuatuoGPT-\\o1-7B~\cite{huatuo-gpt}}} & w/o & 30.05 & 44.45 & 0.30 & 18.56 & 73.63 & 125.79 & 0.24 & 40.26 \\
 & w/ & \textbf{27.85} & \textbf{41.33} & \textbf{0.39} & \textbf{17.32} & \textbf{70.04} & \textbf{118.77} & \textbf{0.35} & \textbf{39.50} \\
\bottomrule
\end{tabular*}
\endgroup
\end{table}

\begin{table}[t]
\centering
\caption{Ablation study on adaptive prior weighting.}
\label{tab:adaptive_prior}
\begingroup
\fontsize{7}{8.5}\selectfont
\setlength{\tabcolsep}{0pt}
\setlength{\arrayrulewidth}{0.3pt}
\renewcommand{\arraystretch}{1.15}
\setlength{\aboverulesep}{1.5pt}
\setlength{\belowrulesep}{2pt}
\newcommand{\adaptmetricwidth}{\dimexpr(\columnwidth-58pt-\arrayrulewidth)/8\relax}
\begin{tabular}{@{}>{\raggedright\arraybackslash}m{43pt}>{\centering\arraybackslash}m{15pt}*{4}{>{\centering\arraybackslash}m{\adaptmetricwidth}}|*{4}{>{\centering\arraybackslash}m{\adaptmetricwidth}}@{}}
\toprule
\multirow{2}{*}{Model} & \multirow{2}{*}{Adapt.}
& \multicolumn{4}{c|}{\textbf{MMSDP}~\cite{xiuwen}}
& \multicolumn{4}{c}{\textbf{EPIC}~\cite{mover}} \\
\cmidrule(lr){3-6}\cmidrule(l){7-10}
& & MAE & RMSE & $R^2$ & MAPE
& MAE & RMSE & $R^2$ & MAPE \\
\midrule
& \texttimes
& 29.68 & 45.04 & 0.27 & 18.22
& 71.19 & 118.80 & 0.35 & 40.91 \\
\rowcolor{black!8}
\multirow{-2}{*}{\cellcolor{white}\shortstack[l]{DeepSeek-\\R1-8B~\cite{deepseek}}}
& \checkmark
& \textbf{29.16} & \textbf{44.66} & \textbf{0.28} & \textbf{17.97}
& \textbf{70.11} & \textbf{118.10} & \textbf{0.36} & \textbf{40.63} \\
\midrule[0.3pt]
& \texttimes
& 28.17 & 41.77 & 0.36 & 17.40
& 70.62 & 119.15 & 0.34 & 39.60 \\
\rowcolor{black!8}
\multirow{-2}{*}{\cellcolor{white}\shortstack[l]{HuatuoGPT-\\o1-7B~\cite{huatuo-gpt}}}
& \checkmark
& \textbf{27.85} & \textbf{41.33} & \textbf{0.39} & \textbf{17.32}
& \textbf{70.04} & \textbf{118.77} & \textbf{0.35} & \textbf{39.50} \\
\bottomrule
\end{tabular}
\endgroup
\end{table}

\begin{table}[t]
\centering
\caption{Ablation of Prior Knowledge in the Prompt.}
\label{tab:prior_ablation}
\begingroup
\fontsize{7.5}{9}\selectfont
\setlength{\tabcolsep}{0.6pt}
\setlength{\arrayrulewidth}{0.3pt}
\renewcommand{\arraystretch}{1.2}
\setlength{\aboverulesep}{1.5pt}
\setlength{\belowrulesep}{2pt}
\begin{tabular*}{\columnwidth}{@{\extracolsep{\fill}}lc*{4}{c}|*{4}{c}@{}}
\toprule
\multirow{2}{*}{Model} & \multirow{2}{*}{Prior}
 & \multicolumn{4}{c|}{\textbf{MMSDP}~\cite{xiuwen}}
 & \multicolumn{4}{c}{\textbf{EPIC}~\cite{mover}} \\
\cmidrule(lr){3-6}\cmidrule(l){7-10}
 & & MAE & RMSE & $R^2$ & MAPE & MAE & RMSE & $R^2$ & MAPE \\
\midrule
\multirow{2}{*}{\shortstack[l]{DeepSeek-\\R1-8B~\cite{deepseek}}} & w/o & 29.74 & 45.18 & 0.26 & 18.28 & 71.50 & 119.03 & 0.34 & 40.94 \\
 & w/ & \textbf{29.16} & \textbf{44.66} & \textbf{0.28} & \textbf{17.97} & \textbf{70.11} & \textbf{118.10} & \textbf{0.36} & \textbf{40.63} \\
\midrule[0.3pt]
\multirow{2}{*}{\shortstack[l]{HuatuoGPT-\\o1-7B~\cite{huatuo-gpt}}} & w/o & 28.23 & 41.93 & 0.36 & 17.44 & 70.54 & 119.07 & \textbf{0.35} & 39.57 \\
 & w/ & \textbf{27.85} & \textbf{41.33} & \textbf{0.39} & \textbf{17.32} & \textbf{70.04} & \textbf{118.77} & \textbf{0.35} & \textbf{39.50} \\
\bottomrule
\end{tabular*}
\endgroup
\end{table}

\begin{table}[h]
\centering
\caption{Ablation study of retrieval post-processing.}
\label{tab:postprocess_ablation}
\begingroup
\fontsize{7.5}{9}\selectfont
\setlength{\tabcolsep}{0.6pt}
\setlength{\arrayrulewidth}{0.3pt}
\renewcommand{\arraystretch}{1.2}
\setlength{\aboverulesep}{1.5pt}
\setlength{\belowrulesep}{2pt}
\begin{tabular*}{\columnwidth}{@{\extracolsep{\fill}}lc*{4}{c}|*{4}{c}@{}}
\toprule
\multirow{2}{*}{Model} & \multirow{2}{*}{Postproc.}
 & \multicolumn{4}{c|}{\textbf{MMSDP}~\cite{xiuwen}}
 & \multicolumn{4}{c}{\textbf{EPIC}~\cite{mover}} \\
\cmidrule(lr){3-6}\cmidrule(l){7-10}
 & & MAE & RMSE & $R^2$ & MAPE & MAE & RMSE & $R^2$ & MAPE \\
\midrule
\multirow{2}{*}{\shortstack[l]{DeepSeek-\\R1-8B~\cite{deepseek}}} & w/o & 30.28 & 46.90 & 0.21 & 19.07 & 71.64 & 120.71 & 0.34 & 41.36 \\
 & w/ & \textbf{29.16} & \textbf{44.66} & \textbf{0.28} & \textbf{17.97} & \textbf{70.11} & \textbf{118.10} & \textbf{0.36} & \textbf{40.63} \\
\midrule[0.3pt]
\multirow{2}{*}{\shortstack[l]{HuatuoGPT-\\o1-7B~\cite{huatuo-gpt}}} & w/o & 28.71 & 42.92 & 0.35 & 17.84 & 71.47 & 124.51 & 0.31 & 39.65 \\
 & w/ & \textbf{27.85} & \textbf{41.33} & \textbf{0.39} & \textbf{17.32} & \textbf{70.04} & \textbf{118.77} & \textbf{0.35} & \textbf{39.50} \\
\bottomrule
\end{tabular*}
\endgroup
\end{table}

\begin{table}[t]
\centering
\caption{Ablation study on the candidate-pool expansion scale in retrieval post-processing on MMSDP.}
\label{tab:postprocess_scale}
\begingroup
\fontsize{7}{8.5}\selectfont
\setlength{\tabcolsep}{0pt}
\setlength{\arrayrulewidth}{0.3pt}
\renewcommand{\arraystretch}{1.15}
\setlength{\aboverulesep}{1.5pt}
\setlength{\belowrulesep}{2pt}
\newcommand{\postmetricwidth}{\dimexpr(\columnwidth-67pt-\arrayrulewidth)/4\relax}
\begin{tabular}{@{}>{\raggedright\arraybackslash}m{52pt}>{\centering\arraybackslash}m{15pt}*{4}{>{\centering\arraybackslash}m{\postmetricwidth}}@{}}
\toprule
Model & \shortstack[c]{\textbf{N $\times$}\\Top-$K$} & MAE $\downarrow$ & RMSE $\downarrow$ & $R^2$$\uparrow$ & MAPE (\%) $\downarrow$\\
\midrule
& 1  & 30.03 & 45.48 & 0.25 & 18.16 \\
& 5  & 29.71 & 45.62 & 0.26 & 18.09 \\
\rowcolor{black!8}
\cellcolor{white} & 10 & \textbf{29.16} & \underline{44.66} & \textbf{0.28} & \textbf{17.97} \\
\multirow{-4}{*}{\shortstack[l]{DeepSeek-\\R1-8B~\cite{deepseek}}}
& 15 & \underline{29.30} & \textbf{44.62} & \underline{0.27} & \underline{17.99} \\
\midrule[0.3pt]
& 1  & 28.41 & 42.54 & 0.36 & \textbf{17.29} \\
& 5  & 28.07 & 42.00 & 0.37 & \underline{17.44} \\
\rowcolor{black!8}
\cellcolor{white} & 10 & \textbf{27.85} & \textbf{41.33} & \textbf{0.39} & \underline{17.32} \\
\multirow{-4}{*}{\shortstack[l]{HuatuoGPT-\\o1-7B~\cite{huatuo-gpt}}}
& 15 & \underline{28.09} & \underline{41.62} & \underline{0.37} & 17.56 \\
\bottomrule
\end{tabular}
\endgroup
\end{table}

\begin{table}[t]
\centering
\caption{Ablation study of heterogeneous feature encoding.}
\label{tab:encoding_ablation}
\begingroup
\fontsize{7}{8.5}\selectfont
\setlength{\tabcolsep}{0.6pt}
\setlength{\arrayrulewidth}{0.3pt}
\renewcommand{\arraystretch}{1.2}
\setlength{\aboverulesep}{1.5pt}
\setlength{\belowrulesep}{2pt}
\begin{tabular*}{\columnwidth}{@{\extracolsep{\fill}}ll|*{4}{c}|*{4}{c}@{}}
\toprule
\multirow{2}{*}{Model} & \multirow{2}{*}{Encoding}
 & \multicolumn{4}{c|}{\textbf{MMSDP}~\cite{xiuwen}}
 & \multicolumn{4}{c}{\textbf{EPIC}~\cite{mover}} \\
\cmidrule(lr){3-6}\cmidrule(l){7-10}
 & & MAE & RMSE & $R^2$ & MAPE & MAE & RMSE & $R^2$ & MAPE \\
\midrule
\multirow{2}{*}{\shortstack[l]{DeepSeek-\\R1-8B~\cite{deepseek}}} & BERT only & 32.68 & 49.42 & 0.12 & 20.17 & 74.02 & 125.28 & 0.27 & 43.21 \\
 & Heterogeneous & \textbf{29.16} & \textbf{44.66} & \textbf{0.28} & \textbf{17.97} & \textbf{70.11} & \textbf{118.10} & \textbf{0.36} & \textbf{40.63} \\
\midrule[0.3pt]
\multirow{2}{*}{\shortstack[l]{HuatuoGPT-\\o1-7B~\cite{huatuo-gpt}}} & BERT only & 30.59 & 44.53 & 0.31 & 18.95 & 72.54 & 123.61 & 0.31 & 42.16 \\
 & Heterogeneous & \textbf{27.85} & \textbf{41.33} & \textbf{0.39} & \textbf{17.32} & \textbf{70.04} & \textbf{118.77} & \textbf{0.35} & \textbf{39.50} \\
\bottomrule
\end{tabular*}
\endgroup
\end{table}

\begin{table}[t]
\centering
\caption{Comparison of BERT encoders for PCA-based retrieval followed by KNN-mean aggregation.}
\label{tab:bert_encoder_ablation}
\begingroup
\fontsize{7}{8.5}\selectfont
\setlength{\tabcolsep}{0pt}
\setlength{\arrayrulewidth}{0.3pt}
\renewcommand{\arraystretch}{1.15}
\setlength{\aboverulesep}{1.5pt}
\setlength{\belowrulesep}{2pt}
\newcommand{\encmetricwidth}{\dimexpr(\columnwidth-121pt-\arrayrulewidth)/4\relax}
\begin{tabular}{@{}>{\raggedright\arraybackslash}m{40pt}>{\raggedright\arraybackslash}m{81pt}*{4}{>{\centering\arraybackslash}m{\encmetricwidth}}@{}}
\toprule
\textbf{Dataset} & \textbf{Encoder} & \textbf{MAE} & \textbf{RMSE} & \textbf{$R^2$} & \textbf{MAPE (\%)} \\
\midrule
\rowcolor{black!8}
\cellcolor{white} & Bert-base-chinese
& \textbf{29.57} & \textbf{45.27} & \textbf{0.27} & \textbf{18.26} \\
& ModernBERT-base-chinese
& 32.02 & 48.01 & 0.17 & 19.65 \\
\multirow{-3}{*}{\shortstack[l]{MMSDP\\\cite{xiuwen}}}
& MedBERT-chinese
& 31.64 & 47.13 & 0.20 & 19.36 \\
\midrule[0.3pt]
& Bert-base-uncased
& 78.70 & 121.75 & 0.32 & 49.91 \\
& ModernBERT-base
& 73.48 & 119.61 & 0.34 & 43.69 \\
& MedBERT-base
& 73.21 & 119.08 & 0.35 & 42.73 \\
\rowcolor{black!8}
\multirow{-4}{*}{\cellcolor{white}\shortstack[l]{EPIC~\cite{mover}}}
& Clinical\_ModernBERT
& \textbf{71.02} & \textbf{117.45} & \textbf{0.36} & \textbf{38.99} \\
\bottomrule
\end{tabular}
\endgroup
\end{table}

\begin{table}[t]
\centering
\caption{Ablation study on effect of the number of prediction rounds.}
\label{tab:prediction_rounds}
\begingroup
\fontsize{7}{8.5}\selectfont
\setlength{\tabcolsep}{0pt}
\setlength{\arrayrulewidth}{0.3pt}
\renewcommand{\arraystretch}{1.15}
\setlength{\aboverulesep}{1.5pt}
\setlength{\belowrulesep}{2pt}
\newcommand{\roundmetricwidth}{\dimexpr(\columnwidth-46pt-\arrayrulewidth)/10\relax}
\begin{tabular}{@{}>{\raggedright\arraybackslash}m{37pt}>{\centering\arraybackslash}m{9pt}*{5}{>{\centering\arraybackslash}m{\roundmetricwidth}}|*{5}{>{\centering\arraybackslash}m{\roundmetricwidth}}@{}}
\toprule
\multirow{2}{*}{Model} & \multirow{2}{*}{$n$}
 & \multicolumn{5}{c|}{\textbf{MMSDP}~\cite{xiuwen}}
 & \multicolumn{5}{c}{\textbf{EPIC}~\cite{mover}} \\
\cmidrule(lr){3-7}\cmidrule(l){8-12}
 & & MAE & RMSE & $R^2$ & MAPE & Time & MAE & RMSE & $R^2$ & MAPE & Time \\
\midrule
 & 1 & 29.25 & \underline{44.60} & \textbf{0.29} & 18.19 & \textbf{1.13} & 71.96 & 123.35 & 0.30 & 41.06 & \textbf{0.63} \\
 & 3 & 29.17 & \textbf{44.56} & \textbf{0.29} & 18.04 & \underline{2.29} & 70.33 & 119.31 & 0.35 & 40.53 & \underline{1.74} \\
\rowcolor{black!8}
\cellcolor{white} & 5 & 29.16 & 44.66 & \underline{0.28} & \underline{17.97} & 3.02 & 70.11 & 118.10 & \underline{0.36} & 40.63 & 2.74 \\
 & 7 & \underline{29.15} & 44.71 & \underline{0.28} & 18.01 & 7.61 & \textbf{69.73} & \underline{117.77} & \underline{0.36} & \textbf{40.41} & 4.36 \\
\multirow{-5}{*}{\shortstack[l]{DeepSeek-\\R1-8B~\cite{deepseek}}} & 9 & \textbf{29.13} & 44.74 & \underline{0.28} & \textbf{17.95} & 8.91 & \underline{69.80} & \textbf{117.22} & \textbf{0.37} & \underline{40.50} & 5.12 \\
\midrule[0.3pt]
 & 1 & 28.86 & 43.25 & 0.33 & 17.86 & \textbf{0.95} & 72.40 & 121.16 & 0.33 & 42.86 & \textbf{0.56} \\
 & 3 & 28.65 & 42.74 & 0.35 & 17.76 & \underline{2.77} & 71.41 & 120.20 & \underline{0.34} & 41.19 & \underline{1.11} \\
\rowcolor{black!8}
\cellcolor{white} & 5 & 27.85 & 41.33 & \underline{0.39} & 17.32 & 4.69 & \textbf{70.04} & \textbf{118.77} & \textbf{0.35} & \textbf{39.50} & 2.63 \\
 & 7 & \underline{27.75} & \underline{41.31} & \underline{0.39} & \underline{17.26} & 4.77 & 70.12 & \underline{118.84} & \textbf{0.35} & 40.14 & 3.36 \\
\multirow{-5}{*}{\shortstack[l]{HuatuoGPT-\\o1-7B~\cite{huatuo-gpt}}} & 9 & \textbf{27.70} & \textbf{41.03} & \textbf{0.40} & \textbf{17.16} & 8.32 & \underline{70.05} & 119.32 & \textbf{0.35} & \underline{39.82} & 3.91 \\
\bottomrule
\end{tabular}
\endgroup
\end{table}

\subsection{\textbf{Case Study}}\label{subsec52}
To provide an intuitive understanding of how PREBA operates in practice, we present a representative case from the in-hospital dataset using Qwen3-8B as the backbone LLM. 
As shown in Fig.~\ref{case_study}, the query corresponds to a 57-year-old male patient scheduled for elective total thyroidectomy in the Department of Breast and Thyroid Surgery, with an ASA grade of II and planned intravenous anesthesia. PREBA selects historical cases that are highly consistent with the query along key clinical dimensions, including department, surgery type and level, ASA grade, anesthesia modality, and a similar age range. The observed durations of these retrieved surgeries are tightly clustered between 115 and 150 minutes, yielding a reference range of [115, 155] minutes and a 95\% confidence interval of (121.16, 146.34). Conditioned on this evidence-based context and population-level priors, PREBA outputs a prediction of 130 minutes, which falls well within the retrieved confidence interval and is close to the ground-truth duration of 120 minutes. This case exemplifies how PREBA retrieves clinically coherent exemplars and combines them with statistical priors to produce calibrated and interpretable surgical duration estimates. 


\subsection{\textbf{Limitation Analysis}}\label{subsec52}

Fig.~\ref{fig:failure} examines two scenarios in which PREBA's core mechanisms may be less effective. Panel (a) shows that Bayesian averaging exhibits a consistent crossover from helping to hurting as retrieved evidence becomes less reliable. Across four datasets, it reduces error by up to 8.4\% when evidence closely matches the truth, but this benefit shrinks and turns into an increase in error of up to 1.6\% once evidence deviates most. This occurs when a query case deviates sharply from its own matched historical cohort: the resulting prior no longer reflects the case's true course, and calibrating toward it can increase rather than reduce the error.

Panel (b) examines the cases with the largest prediction errors on each dataset. Among these cases, retrieval similarity remains at or above the population median in 20.4\%-34.0\% across the four datasets, indicating that many retrieved cases appear similar without being truly representative of the query. 
This may follow from PCA learning a single set of weights across the entire training population and applying it uniformly to every query, without capturing which features matter most for a specific type of surgery. As a result, retrieval can select cases that look similar overall yet differ on the one feature that actually determines the query's duration.

\subsection{\textbf{Ablation Study}}\label{subsec52}
To assess the individual contributions of the main components and key design choices in PREBA, we conduct a series of ablation studies.

\subsubsection{Effect of Top-$K$ Retrieval}
\label{topk_ablation}

We investigate the effect of the number of retrieved historical cases, $K$, on prediction accuracy and inference time. As shown in Table~\ref{tab:topk_ablation}, performance generally improves as $K$ increases from $2$ to $8$, whereas further increasing $K$ to $10$ yields limited or inconsistent gains. 
For DeepSeek-R1-8B, \(K=8\) yields the lowest MAE on MMSDP, whereas \(K=10\) gives the lowest MAE on EPIC. 
For HuatuoGPT-o1-7B, K = 8 gives the lowest MAE on EPIC and is within 0.05 minutes of the best value (K = 6) on MMSDP, attaining the best overall trade-off across both datasets.
These results suggest that retrieving too few cases provides insufficient clinical context, whereas a larger $K$ may introduce less relevant historical cases and increase inference cost. 
This confirms $K=8$ as the default retrieval size in PREBA.

\subsubsection{Ablation Study on $n$-Round Prediction}
\label{round_ablation}
To examine the effect of the number of prediction rounds, $n$, on predictive accuracy and inference cost, we vary $n$ from $1$ to $9$ while keeping all other settings fixed. As shown in Table~\ref{tab:prediction_rounds}, increasing $n$ from $1$ to $5$ generally improves performance. For DeepSeek-R1-8B, MAE decreases from $29.25$ to $29.16$ minutes on MMSDP and from $71.96$ to $70.11$ minutes on EPIC; for HuatuoGPT-o1-7B, the corresponding MAE decreases from $28.86$ to $27.85$ minutes and from $72.40$ to $70.04$ minutes, respectively. Beyond $n=5$, performance improvements become negligible or inconsistent, whereas inference cost increases substantially. For example, on MMSDP with DeepSeek-R1-8B, increasing $n$ from $5$ to $9$ reduces MAE by only $0.03$ minutes ($0.1\%$), while increasing inference time from $3.02$ to $8.91$ seconds ($195\%$). 
This confirms $n=5$ as the default setting.

\subsubsection{Heterogeneous Feature Encoding Ablation}
We conduct an ablation study to evaluate the impact of feature encoding strategies on model performance. Specifically, we compare a BERT-only representation with the proposed heterogeneous encoder, which applies type-specific encoders to heterogeneous clinical features. As shown in Table~\ref{tab:encoding_ablation}, the heterogeneous encoder consistently improves performance across both LLM backbones and datasets. For DeepSeek-R1-8B, it reduces MAE from $32.68$ to $29.16$ minutes on MMSDP and from $74.02$ to $70.11$ minutes on EPIC. Similar improvements are observed for HuatuoGPT-o1-7B, with MAE decreasing from $30.59$ to $27.85$ minutes on MMSDP and from $72.54$ to $70.04$ minutes on EPIC. These results show that type-specific encoding of heterogeneous clinical features yields a more effective retrieval representation than a BERT-only representation.

\subsubsection{Comparison of BERT Encoders}
\label{encoder_ablation}
This study compares BERT encoders for representing clinical textual features in the retrieval stage. 
We evaluate the encoders under PCA-based retrieval with KNN-mean prediction. Table~\ref{tab:bert_encoder_ablation} summarizes the results on MMSDP and EPIC. On MMSDP, Bert-base-chinese achieves the best overall performance, with an MAE of $29.57$ minutes and an RMSE of $45.27$ minutes. On EPIC, Clinical\_ModernBERT performs best across all metrics, attaining an MAE of $71.02$ minutes, an RMSE of $117.45$ minutes, an $R^2$ of $0.36$, and a MAPE of $38.99\%$. Based on these results, we use Bert-base-chinese for the Chinese-language MMSDP and In-hospital datasets and Clinical\_ModernBERT for the English-language EPIC dataset in PREBA.

\begin{figure*}[t]
\centering
\centerline{\includegraphics[width=1\linewidth]{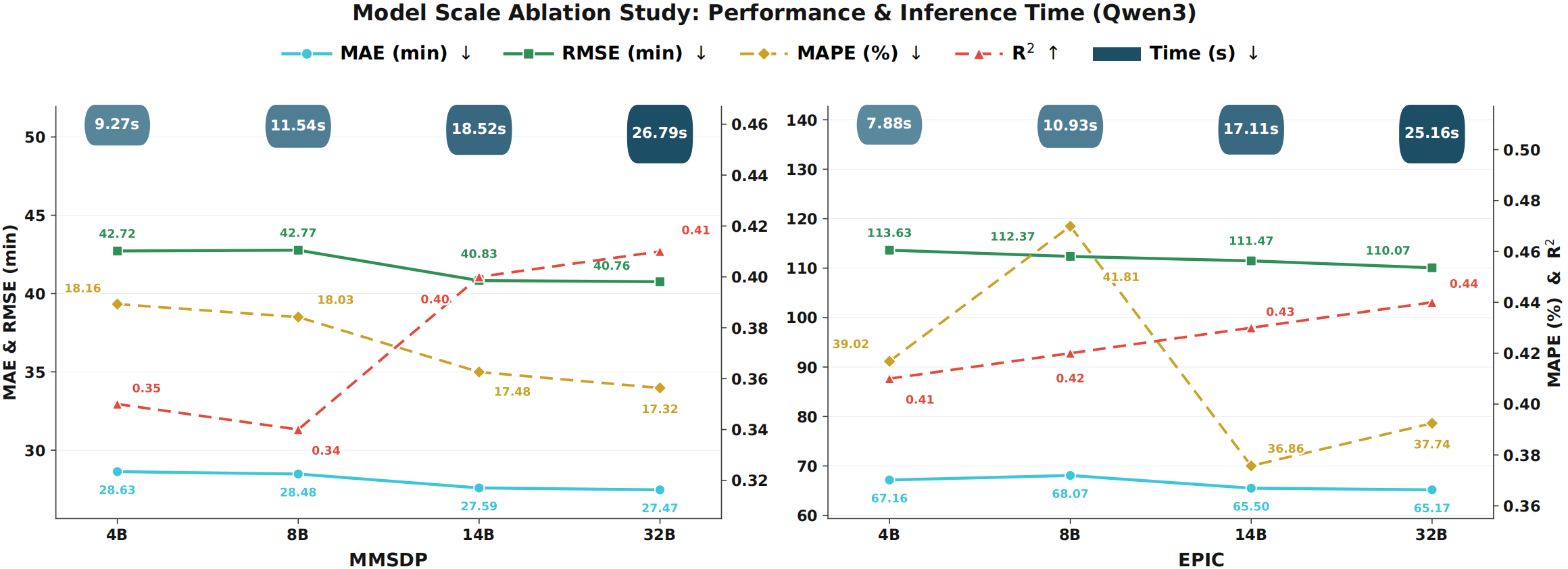}}
\caption{Effect of Qwen3 model scale on prediction performance. 
}
\label{model_scale_ablation}
\end{figure*}

\subsubsection{Ablation on PCA-Weighted Retrieval}
We replace PCA-derived retrieval weights with uniform weights while retaining all remaining components of PREBA. Table~\ref{tab:pca_ablation} shows that PCA weighting improves performance consistently across both LLM backbones and datasets. With DeepSeek-R1-8B, the MAE decreases from $31.79$ to $29.16$ minutes on MMSDP and from $73.68$ to $70.11$ minutes on EPIC. The same effect is observed for HuatuoGPT-o1-7B, for which PCA weighting reduces MAE from $30.05$ to $27.85$ minutes on MMSDP and from $73.63$ to $70.04$ minutes on EPIC. Thus, compared with uniform weighting, PCA weighting enables the retrieval stage to differentiate dimensions within the heterogeneous clinical representation, resulting in more informative historical evidence for LLM-based duration prediction.

\subsubsection{Adaptive Prior Weighting}
\label{adaptive_prior_ablation}
This study evaluates whether prior strength should vary with the amount of matched historical evidence. Let $m$ denote the number of cases in the matched clinical stratum. We assign the prior weight as $0.6$ for $m<50$, $0.9$ for $50 \leq m \leq 100$, and $1.2$ for $m>100$, such that prior information receives greater influence when it is supported by a larger matched cohort. Table~\ref{tab:adaptive_prior} shows that adaptive weighting consistently improves performance across both backbones and datasets. For DeepSeek-R1-8B, it reduces MAE from $29.68$ to $29.16$ minutes on MMSDP and from $71.19$ to $70.11$ minutes on EPIC. Similar improvements are observed for HuatuoGPT-o1-7B, where MAE decreases from $28.17$ to $27.85$ minutes on MMSDP and from $70.62$ to $70.04$ minutes on EPIC. These results support calibrating the contribution of statistical priors according to the amount of available matched evidence.

\subsubsection{Ablation on Statistical Prior Knowledge}
We evaluate the contribution of clinical statistical priors by removing them from the prompt while preserving the remaining PREBA pipeline. The priors summarize historical duration statistics within matched clinical strata defined by key attributes, such as department, planned surgery name. As reported in Table~\ref{tab:prior_ablation}, their inclusion improves prediction accuracy across both datasets and LLM backbones. On MMSDP, adding prior knowledge reduces the MAE of DeepSeek-R1-8B from $29.74$ to $29.16$ minutes and that of HuatuoGPT-o1-7B from $28.23$ to $27.85$ minutes. On EPIC, the corresponding MAEs decrease from $71.50$ to $70.11$ minutes and from $70.54$ to $70.04$ minutes, respectively. These results show that matched clinical statistics provide complementary institution-specific evidence that improves the reliability of LLM-based duration prediction.

\subsubsection{Retrieval Post-processing and the Candidate-Pool Expansion Factor $N$}
\label{postprocess_ablation}
This study evaluates the effect of retrieval post-processing and the candidate-pool expansion factor $N$ on model performance. 
Specifically, an initial pool of $N \times K$ candidates is retrieved and subsequently refined through clinical consistency checks and outlier removal to retain the final $K$ references. Table~\ref{tab:postprocess_ablation} shows that this refinement improves prediction accuracy across both backbones and datasets. For instance, for HuatuoGPT-o1-7B, post-processing reduces MAE from $28.71$ to $27.85$ minutes on MMSDP and from $71.47$ to $70.04$ minutes on EPIC.
We further vary the expansion scale on MMSDP. As reported in Table~\ref{tab:postprocess_scale}, $N=10$ yields the strongest overall performance for both backbones. 
This result suggests that a moderately expanded candidate pool balances diversity against the inclusion of less relevant cases.

\subsubsection{Effect of LLM Scale}
We further investigate how the performance of PREBA varies across LLM backbones of different scales. Qwen3 models with 4B, 8B, 14B, and 32B parameters are evaluated on MMSDP and EPIC. As summarized in Fig.~\ref{model_scale_ablation}, performance generally improves as the backbone scale increases. On MMSDP, the 32B model achieves the best results in terms of MAE ($27.47$ min), RMSE ($40.76$ min), $R^2$ ($0.41$), and MAPE ($17.32\%$). On EPIC, it obtains the lowest MAE ($65.17$ min) and RMSE ($110.07$ min), together with the highest $R^2$ ($0.44$), while the 14B model attains the lowest MAPE ($36.86\%$). The improved accuracy is accompanied by a substantial increase in inference time as the backbone scales from 4B to 32B. Overall, the results demonstrate that PREBA can effectively leverage larger LLM backbones, while practical deployment requires balancing predictive accuracy against inference efficiency.


\section{Conclusion}
We propose PREBA, a retrieval-augmented framework that grounds LLM reasoning in institution-specific evidence for surgical duration prediction without task-specific fine-tuning. It combines PCA-weighted retrieval of clinically relevant historical cases with Bayesian averaging of multi-path LLM predictions against population-level priors. Across four real-world datasets spanning different institutions and languages, PREBA consistently outperforms zero-shot inference and matches supervised machine learning, while offering greater interpretability through retrieved exemplars and explicit priors. Future work will explore multimodal integration of intraoperative video and physiological monitoring data, and continual learning for changing clinical environments.
\label{sec7}


\section*{References}


\begin{thebibliography}{10}
\providecommand{\url}[1]{#1}
\csname url@samestyle\endcsname
\providecommand{\newblock}{\relax}
\providecommand{\bibinfo}[2]{#2}
\providecommand{\BIBentrySTDinterwordspacing}{\spaceskip=0pt\relax}
\providecommand{\BIBentryALTinterwordstretchfactor}{4}
\providecommand{\BIBentryALTinterwordspacing}{\spaceskip=\fontdimen2\font plus
\BIBentryALTinterwordstretchfactor\fontdimen3\font minus \fontdimen4\font\relax}
\providecommand{\BIBforeignlanguage}[2]{{%
\expandafter\ifx\csname l@#1\endcsname\relax
\typeout{** WARNING: IEEEtran.bst: No hyphenation pattern has been}%
\typeout{** loaded for the language `#1'. Using the pattern for}%
\typeout{** the default language instead.}%
\else
\language=\csname l@#1\endcsname
\fi
#2}}
\providecommand{\BIBdecl}{\relax}
\BIBdecl

\bibitem{sdp_manage}
F.~Dexter, R.~H. Epstein, R.~D. Traub, Y.~Xiao, and D.~C. Warltier, ``Making management decisions on the day of surgery based on operating room efficiency and patient waiting times,'' \emph{Anesthesiology}, vol. 101, no.~6, pp. 1444--1453, 2004.

\bibitem{sdp_manage1}
D.~M. Laskin, A.~O. Abubaker, and R.~A. Strauss, ``Accuracy of predicting the duration of a surgical operation,'' \emph{Journal of Oral and Maxillofacial Surgery}, vol.~71, no.~2, pp. 446--447, 2013.

\bibitem{ml_2}
K.~Wang, L.~Z. Yan, W.~Z. Li, C.~Jiang, N.~N. Wang, Q.~Zheng, N.~G. Dong, and J.~W. Shi, ``Comparison of four machine learning techniques for prediction of intensive care unit length of stay in heart transplantation patients,'' \emph{Frontiers in Cardiovascular Medicine}, vol.~9, p. 863642, 2022.

\bibitem{svr}
O.~Martinez, C.~Martinez, C.~A. Parra, S.~Rugeles, and D.~R. Suarez, ``Machine learning for surgical time prediction,'' \emph{Computer Methods and Programs in Biomedicine}, vol. 208, p. 106220, 2021.

\bibitem{sdp_manage2}
D.~R. Roque, K.~Robison, C.~A. Raker, G.~G. Wharton, and G.~N. Frishman, ``The accuracy of surgeons' provided estimates for the duration of hysterectomies: a pilot study,'' \emph{Journal of minimally invasive gynecology}, vol.~22, no.~1, pp. 57--65, 2015.

\bibitem{azriel2024surgery}
D.~Azriel, Y.~Rinott, O.~Tal, B.~Abbou, and N.~Rappoport, ``Surgery duration prediction using multi-task feature selection,'' \emph{IEEE Journal of Biomedical and Health Informatics}, vol.~28, no.~7, pp. 4216--4223, 2024.

\bibitem{nn_1}
Y.~Jiao, B.~Xue, C.~Lu, M.~S. Avidan, and T.~Kannampallil, ``Continuous real-time prediction of surgical case duration using a modular artificial neural network,'' \emph{British journal of anaesthesia}, vol. 128, no.~5, pp. 829--837, 2022.

\bibitem{ml_4}
V.~Riahi, H.~Hassanzadeh, S.~Khanna, J.~Boyle, F.~Syed, B.~Biki, E.~Borkwood, and L.~Sweeney, ``Improving preoperative prediction of surgery duration,'' \emph{BMC Health Services Research}, vol.~23, no.~1, p. 1343, 2023.

\bibitem{Decision_Tree}
L.~Ying \emph{et~al.}, ``Decision tree methods: applications for classification and prediction,'' \emph{Shanghai archives of psychiatry}, vol.~27, no.~2, p. 130, 2015.

\bibitem{XGBoost}
M.~Kwong, M.~Noorchenarboo, K.~Grolinger, J.~Hawel, C.~M. Schlachta, and A.~Elnahas, ``Optimizing surgical efficiency: predicting case duration of common general surgery procedures using machine learning,'' \emph{Surgical Endoscopy}, pp. 1--8, 2025.

\bibitem{nn2}
A.~Ramamurthi, B.~Neupane, P.~Deshpande, R.~Hanson, K.~R. Brown, K.~K. Christians, D.~B. Evans, and A.~N. Kothari, ``Development and validation of an artificial intelligence system for surgical case length prediction,'' \emph{Surgery}, vol. 179, p. 108942, 2025.

\bibitem{muep}
K.~Li, B.~Yu, Q.~Zheng, Y.~Zhan, Y.~Zhang, T.~Zhang, Y.~Yang, Y.~Chen, L.~Sun, Q.~Cao \emph{et~al.}, ``Muep: A multimodal benchmark for embodied planning with foundation models [c],'' in \emph{Intemational Joint Conferences on Artificial Intelligence. IJCAI}, 2024, pp. 129--138.

\bibitem{llm4sdp}
A.~Ramamurthi, B.~Neupane, P.~Deshpande, R.~Hanson, S.~Vegesna, D.~Cray, B.~H. Crotty, M.~Somai, K.~R. Brown, S.~S. Pawar \emph{et~al.}, ``Applying large language models for surgical case length prediction,'' \emph{JAMA surgery}, vol. 160, no.~8, pp. 894--902, 2025.

\bibitem{case_mix}
H.~Elayan, M.~Sperrin, G.~P. Martin, N.~Peek, F.~Braunschweig, J.~Fax{\'e}n, J.~Alfredsson, and D.~A. Jenkins, ``Correcting for case-mix shift when developing clinical prediction models,'' \emph{BMC Medical Research Methodology}, vol.~25, no.~1, p. 186, 2025.

\bibitem{rag_review2}
W.~Fan, Y.~Ding, L.~Ning, S.~Wang, H.~Li, D.~Yin, T.-S. Chua, and Q.~Li, ``A survey on rag meeting llms: Towards retrieval-augmented large language models,'' in \emph{Proceedings of the 30th ACM SIGKDD conference on knowledge discovery and data mining}, 2024, pp. 6491--6501.

\bibitem{fang2021patient}
H.~S.~A. Fang, N.~C. Tan, W.~Y. Tan, R.~W. Oei, M.~L. Lee, and W.~Hsu, ``Patient similarity analytics for explainable clinical risk prediction,'' \emph{BMC medical informatics and decision making}, vol.~21, no.~1, p. 207, 2021.

\bibitem{wan2025reasoning}
G.~Wan, Y.~Wu, J.~Chen, and S.~Li, ``Reasoning aware self-consistency: Leveraging reasoning paths for efficient llm sampling,'' in \emph{Proceedings of the 2025 Conference of the Nations of the Americas Chapter of the Association for Computational Linguistics: Human Language Technologies (Volume 1: Long Papers)}, 2025, pp. 3613--3635.

\bibitem{pca}
A.~Ma{\'c}kiewicz and W.~Ratajczak, ``Principal components analysis (pca),'' \emph{Computers \& Geosciences}, vol.~19, no.~3, pp. 303--342, 1993.

\bibitem{self_consistency}
\BIBentryALTinterwordspacing
X.~Wang, J.~Wei, D.~Schuurmans, Q.~V. Le, E.~H. Chi, S.~Narang, A.~Chowdhery, and D.~Zhou, ``Self-consistency improves chain of thought reasoning in language models,'' in \emph{The Eleventh International Conference on Learning Representations, {ICLR} 2023, Kigali, Rwanda, May 1-5, 2023}.\hskip 1em plus 0.5em minus 0.4em\relax OpenReview.net, 2023. [Online]. Available: \url{https://openreview.net/forum?id=1PL1NIMMrw}
\BIBentrySTDinterwordspacing

\bibitem{bayesian}
T.~M. Fragoso, W.~Bertoli, and F.~Louzada, ``Bayesian model averaging: A systematic review and conceptual classification,'' \emph{International Statistical Review}, vol.~86, no.~1, pp. 1--28, 2018.

\bibitem{qwen3}
A.~Yang, A.~Li, B.~Yang, B.~Zhang, B.~Hui, B.~Zheng, B.~Yu, C.~Gao, C.~Huang, C.~Lv \emph{et~al.}, ``Qwen3 technical report,'' \emph{arXiv preprint arXiv:2505.09388}, 2025.

\bibitem{deepseek}
D.~Guo, D.~Yang, H.~Zhang, J.~Song, R.~Zhang, R.~Xu, Q.~Zhu, S.~Ma, P.~Wang, X.~Bi \emph{et~al.}, ``Deepseek-r1: Incentivizing reasoning capability in llms via reinforcement learning,'' \emph{arXiv preprint arXiv:2501.12948}, 2025.

\bibitem{huatuo-gpt}
J.~Chen, Z.~Cai, K.~Ji, X.~Wang, W.~Liu, R.~Wang, and B.~Wang, ``Towards medical complex reasoning with llms through medical verifiable problems,'' in \emph{Findings of the Association for Computational Linguistics: ACL 2025}, 2025, pp. 14\,552--14\,573.

\bibitem{Bagging}
L.~Breiman, ``Bagging predictors,'' \emph{Machine learning}, vol.~24, no.~2, pp. 123--140, 1996.

\bibitem{QrandForest}
A.~Dean, A.~Meisami, H.~Lam, M.~P. Van~Oyen, C.~Stromblad, and N.~Kastango, ``Quantile regression forests for individualized surgery scheduling,'' \emph{Health Care Management Science}, vol.~25, no.~4, pp. 682--709, 2022.

\bibitem{bert}
W.~Lan, Z.~Tang, M.~Liu, Q.~Chen, W.~Peng, Y.-p.~P. Chen, and Y.~Pan, ``The large language models on biomedical data analysis: a survey,'' \emph{IEEE Journal of Biomedical and Health Informatics}, vol.~29, no.~6, pp. 4486--4497, 2025.

\bibitem{luo2024biomedgpt}
Y.~Luo, J.~Zhang, S.~Fan, K.~Yang, M.~Hong, Y.~Wu, M.~Qiao, and Z.~Nie, ``Biomedgpt: An open multimodal large language model for biomedicine,'' \emph{IEEE journal of biomedical and health informatics}, 2024.

\bibitem{aamodt1994case}
P.~Herrero, P.~Pesl, M.~Reddy, N.~Oliver, P.~Georgiou, and C.~Toumazou, ``Advanced insulin bolus advisor based on run-to-run control and case-based reasoning,'' \emph{IEEE journal of biomedical and health informatics}, vol.~19, no.~3, pp. 1087--1096, 2014.

\bibitem{li2025cross}
K.~Li, Y.~Zhan, H.~Jin, C.~Qi, X.~Lin, and B.~Yu, ``Cross-sample augmented test-time adaptation for personalized intraoperative hypotension prediction,'' in \emph{Proceedings of the AAAI Conference on Artificial Intelligence}, vol.~40, no.~27, 2026, pp. 22\,958--22\,966.

\bibitem{rag_review}
M.~p. p. o. s.~d. Arslan, H.~Ghanem, S.~Munawar, and C.~Cruz, ``A survey on rag with llms,'' \emph{Procedia computer science}, vol. 246, pp. 3781--3790, 2024.

\bibitem{realtime_tmm}
Y.~Kang, X.~Wang, Y.~Xu, and Y.~Wang, ``Multi-strategy rag for disease comorbidity prediction,'' \emph{IEEE Journal of Biomedical and Health Informatics}, 2026.

\bibitem{action}
K.~Li, Q.~Zheng, Y.~Zhan, C.~Zhang, T.~Zhang, X.~Lin, C.~Qi, L.~Li, and D.~Tao, ``Alleviating action hallucination for llm-based embodied agents via inner and outer alignment,'' in \emph{2024 7th International Conference on Pattern Recognition and Artificial Intelligence (PRAI)}.\hskip 1em plus 0.5em minus 0.4em\relax IEEE, 2024, pp. 613--621.

\bibitem{medical_rag}
F.~Neha, D.~Bhati, and D.~K. Shukla, ``Retrieval-augmented generation (rag) in healthcare: A comprehensive review,'' \emph{AI}, vol.~6, no.~9, p. 226, 2025.

\bibitem{guyon2003introduction}
I.~Guyon and A.~Elisseeff, ``An introduction to variable and feature selection,'' \emph{Journal of machine learning research}, vol.~3, no. Mar, pp. 1157--1182, 2003.

\bibitem{altmann2010permutation}
A.~Altmann, L.~Tolo{\c{s}}i, O.~Sander, and T.~Lengauer, ``Permutation importance: a corrected feature importance measure,'' \emph{Bioinformatics}, vol.~26, no.~10, pp. 1340--1347, 2010.

\bibitem{mosca2022shap}
E.~Mosca, F.~Szigeti, S.~Tragianni, D.~Gallagher, and G.~Groh, ``Shap-based explanation methods: a review for nlp interpretability,'' in \emph{Proceedings of the 29th international conference on computational linguistics}, 2022, pp. 4593--4603.

\bibitem{madaan2023self}
A.~Madaan, N.~Tandon, P.~Gupta, S.~Hallinan, L.~Gao, S.~Wiegreffe, U.~Alon, N.~Dziri, S.~Prabhumoye, Y.~Yang \emph{et~al.}, ``Self-refine: Iterative refinement with self-feedback,'' \emph{Advances in neural information processing systems}, vol.~36, pp. 46\,534--46\,594, 2023.

\bibitem{scikit-learn}
F.~Pedregosa, G.~Varoquaux, A.~Gramfort, V.~Michel, B.~Thirion, O.~Grisel, M.~Blondel, P.~Prettenhofer, R.~Weiss, V.~Dubourg \emph{et~al.}, ``Scikit-learn: Machine learning in python,'' \emph{the Journal of machine Learning research}, vol.~12, pp. 2825--2830, 2011.

\bibitem{faiss}
M.~Douze, A.~Guzhva, C.~Deng, J.~Johnson, G.~Szilvasy, P.-E. Mazar{\'e}, M.~Lomeli, L.~Hosseini, and H.~J{\'e}gou, ``The faiss library,'' \emph{IEEE Transactions on Big Data}, 2025.

\bibitem{cosine}
R.~P. Srivastava, ``A new measure of similarity in textual analysis: Vector similarity metric versus cosine similarity metric,'' \emph{Journal of Emerging Technologies in Accounting}, vol.~20, no.~1, pp. 77--90, 2023.

\bibitem{iqr}
X.~Wan, W.~Wang, J.~Liu, and T.~Tong, ``Estimating the sample mean and standard deviation from the sample size, median, range and/or interquartile range,'' \emph{BMC medical research methodology}, vol.~14, no.~1, p. 135, 2014.

\bibitem{vote_med}
H.~Yang, M.~Li, H.~Zhou, Y.~Xiao, Q.~Fang, and R.~Zhang, ``One llm is not enough: Harnessing the power of ensemble learning for medical question answering,'' \emph{medRxiv}, 2023.

\bibitem{xiuwen}
X.~Li, Y.~Zhan, J.~Ni, F.~Cai, H.~Jin, X.~Lin, Y.~Zhang, and D.~Tao, ``Multimodal feature analysis for surgery duration predication,'' in \emph{Proceedings of the 2024 16th International Conference on Bioinformatics and Biomedical Technology}, 2024, pp. 188--195.

\bibitem{mover}
M.~Samad, M.~Angel, J.~Rinehart, Y.~Kanomata, P.~Baldi, and M.~Cannesson, ``Medical informatics operating room vitals and events repository (mover): a public-access operating room database,'' \emph{JAMIA open}, vol.~6, no.~4, p. ooad084, 2023.

\bibitem{inspire}
L.~Lim, H.~Lee, C.-W. Jung, D.~Sim, X.~Borrat, T.~J. Pollard, L.~A. Celi, R.~G. Mark, S.~T. Vistisen, and H.-C. Lee, ``Inspire, a publicly available research dataset for perioperative medicine,'' \emph{Scientific Data}, vol.~11, no.~1, p. 655, 2024.

\bibitem{Gradient_Boosting_Trees}
R.~A. Gabriel, B.~Harjai, S.~Simpson, A.~L. Du, J.~L. Tully, O.~George, and R.~Waterman, ``An ensemble learning approach to improving prediction of case duration for spine surgery: algorithm development and validation,'' \emph{JMIR Perioperative Medicine}, vol.~6, p. e39650, 2023.

\bibitem{randForest}
S.~J. Rigatti, ``Random forest,'' \emph{Journal of insurance medicine}, vol.~47, no.~1, pp. 31--39, 2017.

\bibitem{linear_regression}
E.~R. Edelman, S.~M. Van~Kuijk, A.~E. Hamaekers, M.~J. De~Korte, G.~G. Van~Merode, and W.~F. Buhre, ``Improving the prediction of total surgical procedure time using linear regression modeling,'' \emph{Frontiers in medicine}, vol.~4, p.~85, 2017.

\bibitem{Ridge_Regression}
A.~M.~E. Saleh, M.~Arashi, and B.~G. Kibria, \emph{Theory of ridge regression estimation with applications}.\hskip 1em plus 0.5em minus 0.4em\relax John Wiley \& Sons, 2019.

\bibitem{walters2009estimation}
J.~Walters-Williams and Y.~Li, ``Estimation of mutual information: A survey,'' in \emph{International Conference on Rough Sets and Knowledge Technology}.\hskip 1em plus 0.5em minus 0.4em\relax Springer, 2009, pp. 389--396.

\bibitem{scornet2023trees}
E.~Scornet, ``Trees, forests, and impurity-based variable importance in regression,'' in \emph{Annales de l'Institut Henri Poincare (B) Probabilites et statistiques}, vol.~59, no.~1.\hskip 1em plus 0.5em minus 0.4em\relax Institut Henri Poincar{\'e}, 2023, pp. 21--52.

\end{thebibliography}
\end{document}